\documentclass{article}

\PassOptionsToPackage{numbers,sort&compress}{natbib}
\usepackage[preprint]{neurips_2026}

\usepackage[utf8]{inputenc}
\usepackage[T1]{fontenc}
\usepackage{booktabs}
\usepackage{amsfonts}
\usepackage{amsmath}
\usepackage{amssymb}
\usepackage{amsthm}
\usepackage{nicefrac}
\usepackage{microtype}
\usepackage{xcolor}
\usepackage{mathtools}
\usepackage{enumitem}
\usepackage{algorithm}
\usepackage{algorithmic}
\usepackage[table]{xcolor}
\usepackage{multirow}
\usepackage{tabularx}
\usepackage{wrapfig}
\usepackage{graphicx}
\usepackage{makecell}
\usepackage[most]{tcolorbox}
\usepackage[colorlinks=true,
            citecolor=violet,
            linkcolor=violet,
            urlcolor=violet]{hyperref}\usepackage{url}
\usepackage[nameinlink,noabbrev,capitalise]{cleveref}
\usepackage{caption}

\newcommand{\ours}{\textsc{SegTreeMem}}

\let\Autoref\relax
\DeclareRobustCommand{\Autoref}[1]{\Cref{#1}}

\crefname{figure}{Figure}{Figures}
\Crefname{figure}{Figure}{Figures}
\crefname{table}{Table}{Tables}
\Crefname{table}{Table}{Tables}
\crefname{section}{Section}{Sections}
\Crefname{section}{Section}{Sections}
\crefname{subsection}{Section}{Sections}
\Crefname{subsection}{Section}{Sections}
\crefname{equation}{Equation}{Equations}
\Crefname{equation}{Equation}{Equations}
\crefname{algorithm}{Algorithm}{Algorithms}
\Crefname{algorithm}{Algorithm}{Algorithms}

\theoremstyle{definition}

\title{Temporal Order Matters for Agentic Memory: Segment Trees for Memory Construction and Retrieval}
\author{
\textbf{Yifan Simon Liu}\textsuperscript{1},
\textbf{Liam Gallagher}\textsuperscript{1},
\textbf{Faeze Moradi Kalarde}\textsuperscript{1},\\
\textbf{Jiazhou Liang}\textsuperscript{1},
\textbf{Armin Toroghi}\textsuperscript{1},
\textbf{Scott Sanner}\textsuperscript{1,2}
\\[0.5em]
\textsuperscript{1}University of Toronto
\\
\textsuperscript{2}Vector Institute for Artificial Intelligence
\\
\texttt{yifanliu.liu@mail.utoronto.ca}
}

\newcounter{algobox}

\newcommand{\appref}[1]{\hyperref[#1]{Appendix~\ref*{#1}}}

\usepackage[most]{tcolorbox}
\tcbuselibrary{skins,breakable}
\usepackage{tabularx}

\definecolor{purplebar}{HTML}{6B4E8A}   
\definecolor{purplebg}{HTML}{F5F1F8}    
\definecolor{slatebar}{HTML}{5A6B82}    
\definecolor{slatebg}{HTML}{F0F2F5}     
\definecolor{rulegrey}{HTML}{C8CDD3}    

\newtcolorbox{mem2cell}{%
  enhanced,
  colback=purplebg, colframe=purplebar,
  boxrule=0pt, leftrule=2.5pt,
  sharp corners,
  left=6pt, right=6pt, top=4pt, bottom=4pt,
  before skip=0pt, after skip=0pt,
}
\newtcolorbox{mtcell}{%
  enhanced,
  colback=slatebg, colframe=slatebar,
  boxrule=0pt, leftrule=2.5pt,
  sharp corners,
  left=6pt, right=6pt, top=4pt, bottom=4pt,
  before skip=0pt, after skip=0pt,
}

\newcommand{\compare}[2]{%
  \par\vspace{2pt}%
  \noindent
  \begin{minipage}[t]{0.485\linewidth}#1\end{minipage}%
  \hspace{0.02\linewidth}%
  \begin{minipage}[t]{0.485\linewidth}#2\end{minipage}%
  \par\vspace{6pt}%
}

\begin{document}

\maketitle


\begin{abstract}
Long-horizon conversational agents need to interact with users through evolving events, tasks, and goals, where utterances within a period often center on a shared topic before the conversation shifts to a new context. Such histories can be naturally represented as a temporally ordered segment tree, where sequential utterances compose into higher-level segments.
However, existing tree-based memory systems mostly focus on topically structured hierarchies and often ignore temporal sequencing, leaving two key design questions for temporally ordered memory underexplored: (i) how to update a temporally ordered tree online as new utterances arrive, and (ii) how to exploit the resulting temporal hierarchy along with topical structure during retrieval. 
We introduce Segment Tree Memory (\ours), a memory architecture that represents
conversation history as a Segment Tree over utterances while preserving temporal
order. To support online updates, \ours\ incrementally incorporates new
utterances into the Segment Tree while maintaining temporally contiguous
interaction segments. For retrieval, \ours\ propagates relevance signals through
the tree, enabling the model to integrate information across related parts of
the conversation and identify query-relevant context more effectively. Our experiments show that across three datasets and two LLM backbones, \ours\ improves LLM-judge accuracy
by nearly 20\% over non-temporal tree baselines and other strong memory
baselines. These results support the view that long-horizon conversational
agents benefit from memory indexes that are \emph{both} temporal and hierarchical.

\end{abstract}

\section{Introduction}
\label{sec:introduction}

Long-horizon conversational agents interact with users over multi-turn dialogues
of sequential user and agent utterances~\citep{locomo2024,
longmemeval2024,memoryagentbench2025}. To respond accurately to a query, the
agent must retain and leverage relevant information from earlier utterances. This
necessitates a memory system that supports efficient \textit{online construction}
as utterances arrive to retain the full history, along with a
\textit{retrieval mechanism} that accesses query-relevant information to support
response generation~\citep{generativeagents2023,memgpt2023,mem02025}.

\begin{figure}[t]
  \centering
  \includegraphics[
    width=0.95\textwidth,
    keepaspectratio
  ]{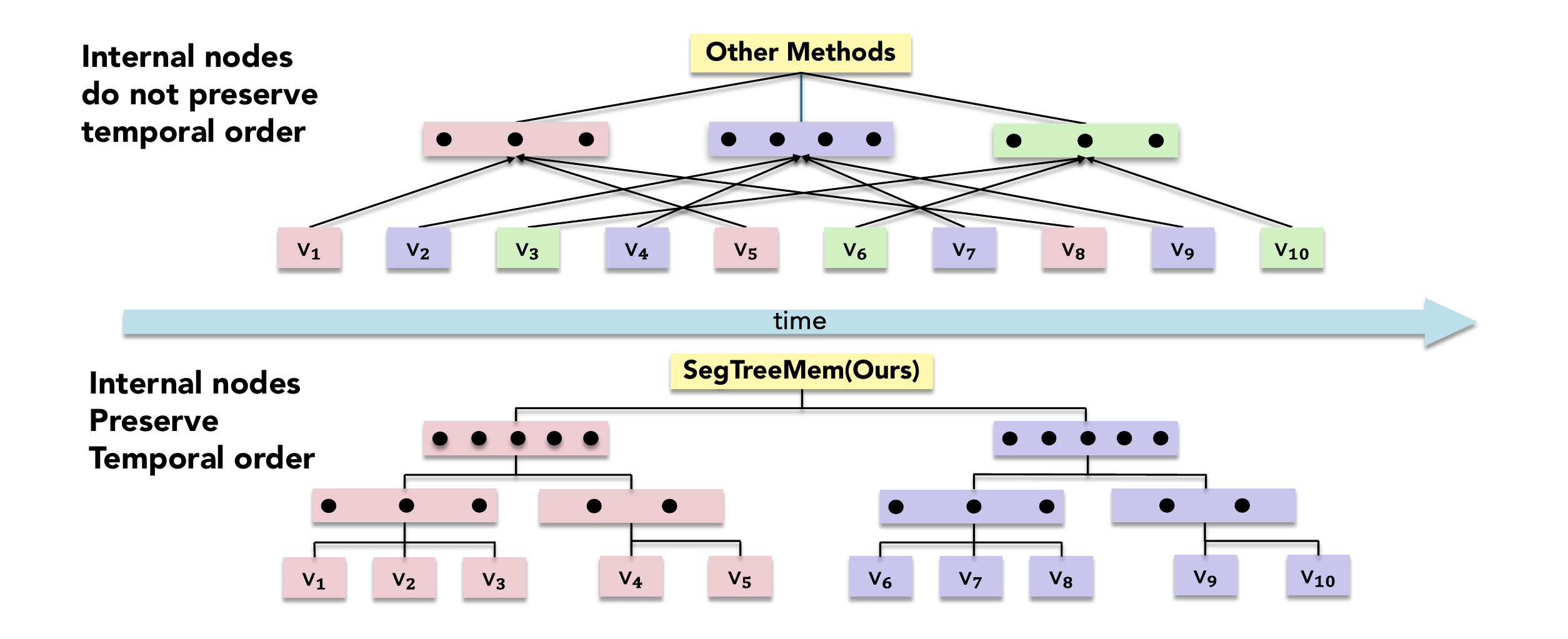}
\caption{Memory tree representations. Semantic trees may group non-consecutive utterances, ignoring temporal order (top), while \ours\ preserves temporal order by assigning each node a contiguous span of utterances (bottom).}
\vspace{-1em}
  \label{fig:teaser}
\end{figure}

In real-world interactions, conversations naturally organize into topically coherent and temporally ordered segments, where consecutive utterances share the same subject before transitioning to a new topic~\citep{galley2003discourse,joty2013topic}. This implies that the relevance of a past utterance to a current query cannot be determined from its content alone, as utterances within a segment are semantically interrelated and their meaning depends on surrounding utterances within the same segment. A memory system that does not account for temporal order risks retrieving semantically similar yet contextually misaligned information. This makes tree-based memory structures a natural choice, as they can organize conversations into hierarchically nested, temporally ordered segments while preserving local contextual dependencies.

However, existing tree-based memory systems tend to focus on topically structured hierarchies and often place less emphasis on temporal sequencing, leaving open two key design questions for temporally ordered tree memory:
    (i) \textit{How should the tree be constructed online to index incoming
    utterances in real-time while preserving topical coherence and temporal order?}
    (ii) \textit{How should retrieval exploit both topical and temporal organization in the resulting tree structure?}

To preserve the temporal order of utterances in memory, we use a Segment Tree~\cite{deberg2008computational}, where each node represents a contiguous span
of utterances, and internal nodes recursively partition their spans into ordered,
non-overlapping sub-spans. Building on the Segment Tree structure, we introduce Segment Tree Memory (\ours\ shown in \Autoref{fig:teaser}). Specifically, our contributions are as follows:



\begin{itemize}
\item \ours\ inserts each new utterance by updating only a small set of nodes
along the rightmost frontier, supporting real-time tree construction while preserving temporal order and coherent node representations (\Autoref{fig:online_updates}).

\item \ours\ employs a structure-aware retrieval mechanism that uses the tree as
a temporal and hierarchical memory index and formulates retrieval as relevance
propagation over the tree via a transition matrix, combining local semantic
matching with hierarchical topical structure and temporal dependencies
(\Autoref{fig:retrieval}).


\item We empirically evaluate \ours\ on three memory datasets and
two LLM backbones, showing consistent accuracy improvements over existing
baselines and demonstrating the benefit of temporal and hierarchical memory indexes for
conversational agents.

\end{itemize}

\section{Problem Definition}
\label{sec:preliminaries}

We consider a conversational agent that interacts with a user through a sequence
of utterances. Let \(x_t\) denote the \(t\)-th utterance. The conversation history after \(t\) utterances is
\(X_t=(x_1,\ldots,x_t)\).

A long-horizon agent maintains an external memory state throughout the
conversation. Let \(M_t\) denote the memory state after processing the first
\(t\) utterances. In this work, we focus on two components of memory: online memory construction and the retrieval from memory, which are described below.

\paragraph{Online Memory Construction.}
When a new utterance \(x_{t+1}\) arrives, the memory state is updated as
\(M_{t+1}=\mathcal U(M_t,x_{t+1})\), where \(\mathcal U\) denotes the online
update operator. An effective \(\mathcal U\) should satisfy:
\begin{enumerate}
    \item[(i)]\textbf{Incremental update:}
    It incorporates \(x_{t+1}\) into \(M_t\) without rebuilding memory from
    scratch.
    \item[(ii)]\textbf{Efficient update scope:}
    Let \(\mathcal C_t\) and \(\Delta_t\) denote the memory units inspected and
    modified during update. For real-time interaction,
    \(|\mathcal C_t|,|\Delta_t|\ll |M_t|\), ideally sublinear in
    \(|M_t|\).
   \item[(iii)]\textbf{Structural and temporal preservation:}
    The updated memory \(M_{t+1}\) should preserve structural organization and
    conversation order. Memory units may be grouped or abstracted, but not reordered, i.e., \(i<j\Rightarrow x_i\) precedes \(x_j\) in \(M_{t+1}\).
\end{enumerate}
\paragraph{Retrieval from Memory.}
Given a user query \(q\), the agent retrieves evidence
\(O=\mathcal R(q,M_t)\) from the current memory state, where \(\mathcal R\) is
the retrieval operator. An effective \(\mathcal R\) should satisfy:
\begin{enumerate}
\item[(i)]\textbf{Topical relevance:}
    Let \(R(q,m)\) denote the topical relevance between query \(q\) and memory
    unit \(m\). Retrieval should favor memory units with high \(R(q,m)\).
\item[(ii)]\textbf{Temporal and structural context:}
    A topical match may not be sufficient alone. Let \(\mathcal N_t(m)\) denote the
    structurally and temporally admissible neighborhood of \(m\) in \(M_t\).
    Retrieval should use topical matches as memory entry points and
    \(\mathcal N_t(m)\) to expand the evidence.
\end{enumerate}

\section{Related Work}
\label{sec:related_work}


\paragraph{Memory construction.}
RAPTOR~\citep{raptor2024} and LATTICE~\citep{lattice2025} build semantic
abstraction trees through clustering. MemWalker~\citep{memwalker2023} preserves
document order by grouping adjacent segments into fixed-size units.
MemTree~\citep{memtree2024} builds a semantic tree by inserting new memories
into the most similar branch. Among these methods, only MemTree explicitly
supports online memory update. However, its similarity-based insertion
prioritizes topical relatedness and does not preserve utterance order in the
tree structure. Conversely, MemWalker preserves temporal order but is designed
for static contexts rather than online updates. \ours\ supports both.

\paragraph{Retrieval from memory.}
Existing retrieval methods vary in whether and how they utilize the tree structure during retrieval. RAPTOR~\citep{raptor2024} and MemTree~\citep{memtree2024} perform \emph{collapsed retrieval}, where all tree nodes are treated as independent candidates.
Although this approach supports topical relevance, it does not explicitly use structural or temporal context during retrieval. MemWalker~\citep{memwalker2023} and LATTICE~\citep{lattice2025} perform \emph{LLM-guided traversal}, where an LLM navigates the tree hierarchy. Although these methods leverage the top-down tree structure, they do not exploit other tree-induced relations or temporal order during retrieval. In contrast, the retrieval mechanism in \ours\ exploits both the tree structure and temporal order to identify additional relevant context.

\definecolor{lightpurple}{RGB}{244,238,255}
\definecolor{headerpurple}{RGB}{226,214,245}
\definecolor{goodgreen}{RGB}{0,128,80}
\definecolor{badred}{RGB}{190,50,50}

\newcommand{\cmark}{\textcolor{goodgreen}{\(\checkmark\)}}
\newcommand{\xmark}{\textcolor{badred}{\(\times\)}}

\begin{table}[t]
\centering
\small
\setlength{\tabcolsep}{4pt}
\renewcommand{\arraystretch}{1.15}
\resizebox{\columnwidth}{!}{%
\begin{tabular}{lccc ccc}
\toprule
\rowcolor{headerpurple}
\cellcolor{white}
&
\multicolumn{3}{c}{\textbf{Construction}} &
\multicolumn{3}{c}{\textbf{Retrieval}} \\
\cmidrule(lr){2-4}\cmidrule(lr){5-7}
\rowcolor{headerpurple}
\cellcolor{white}
& \textbf{Build} & \textbf{Online} & \textbf{Preserves order}
& \textbf{Retrieval} & \textbf{Uses hierarchy} & \textbf{Uses order} \\
\midrule
RAPTOR~\citep{raptor2024} &
Clustering &
\xmark &
\xmark &
Collapsed &
\xmark &
\xmark \\

MemTree~\citep{memtree2024} &
Semantic insertion &
\cmark &
\xmark &
Collapsed &
\xmark &
\xmark \\

MemWalker~\citep{memwalker2023} &
Fixed ordered grouping &
\xmark &
\cmark &
LLM-guided traversal &
\cmark &
\cmark \\

LATTICE~\citep{lattice2025} &
Clustering &
\xmark &
\xmark &
LLM-guided traversal &
\cmark &
\xmark \\

\rowcolor{lightpurple}
\ours\ (Ours) &
Segment Tree &
\cmark &
\cmark &
Relevance propagation &
\cmark &
\cmark \\
\bottomrule
\end{tabular}%
}
\captionsetup{skip=10pt}
\caption{
Comparison of tree-based memory methods through the desiderata in
\Autoref{sec:preliminaries}. Under \textbf{Construction}, \textbf{Online} indicates whether the method defines an online update rule for
incorporating a new utterance into an existing memory state, and
\textbf{Preserves order} indicates whether construction explicitly preserves
conversation order. Under
\textbf{Retrieval}, \textbf{Uses hierarchy} indicates whether retrieval uses the
tree structure rather than treating nodes as independent candidates, and
\textbf{Uses order} indicates whether retrieval can exploit temporal order when selecting evidence.
}
\label{tab:tree_method_comparison}
\end{table}








\section{Methodology}
\label{sec:method}

A Segment Tree~\cite{deberg2008computational} provides a principled data structure that satisfies the desiderata of online memory construction in \Autoref{sec:preliminaries}. It maintains a hierarchical decomposition over contiguous time intervals, where
(i) incorporating \(x_{t+1}\) reduces to attaching a rightmost leaf and updating nodes along a single root-to-leaf path, avoiding global reconstruction;
(ii) the structure restricts inspection and modification to \(O(\log |M_t|)\) nodes under balanced tree;
(iii) each node encodes a contiguous segment, and the left-to-right leaf order follows chronology, preserving temporal consistency.



\subsection{Conversation Segment Tree}
\label{sec:conversation_segment_tree}

Given the temporally ordered utterance sequence \(X=(x_1,\dots,x_T)\), where
\(x_t\) denotes the \(t\)-th utterance, we represent \(X\) as a rooted Segment
Tree \(T=(V,E,\rho)\) to preserve temporal order for later retrieval
(\autoref{fig:teaser}, bottom). Here, \(V\) is the set of nodes, \(E\) is the
set of parent-child edges, and \(\rho\in V\) is the root node.

Each node \(v\in V\) is associated with a contiguous interval of the conversation
history, denoted by \(I(v)=\big[l(v),r(v) \big]\), where \(l(v)\) and \(r(v)\)
are the indices of the first and last utterances in the conversation interval,
respectively, with \(1 \le l(v) \le r(v) \le T\). Thus, \(v\) represents the subsequence
\(\big(x_{l(v)},x_{l(v)+1},\dots,x_{r(v)}\big)\) in the conversation history. The tree comprises three types of nodes.

\paragraph{Root node.} The root node \(\rho\) is associated with the entire
conversation history, i.e., \(I(\rho)=[1,T]\).
\paragraph{Internal nodes.}
Each internal node \(v\) is associated with an interval equal to the union of the
intervals of its child nodes. Let \(\mathsf{ch}(v)=(c_1,\dots,c_m)\) denote the
ordered children of node \(v\), indexed from left to right. These children have
pairwise disjoint and consecutive intervals, i.e.,
\begin{align}
r(c_j)+1=l(c_{j+1}), \qquad 1\le j \le m-1.
\end{align}
Then, the interval associated with node \(v\) is
$
I(v) = \bigcup_{j=1}^{m} \big[l(c_j),r(c_j)\big]  = \big[l(c_1),r(c_m)\big],
$
where the last equality follows from the fact that intervals associated with consecutive child nodes are adjacent and non-overlapping. Therefore, the
interval associated with every internal node is partitioned into consecutive
non-overlapping sub-intervals, each corresponding to a child node.

\paragraph{Leaf nodes.}
Each leaf node represents a single utterance. Thus, for every leaf node
\(v\), \(I(v)=[t,t]\).

\paragraph{Node annotations.}
Each node \(v\) carries an annotation \(A(v)\), which is an LLM-generated summary
of the conversation span \(I(v)\).


\subsection{Online Segment Tree Construction}
\label{sec:online_construction}


In the agent setting, memory must be constructed online as the conversation
unfolds. Given \(X_t=(x_1,\dots,x_t)\), let
\(T_t=(V_t,E_t,\rho_t)\) be the Segment Tree of the first \(t\) utterances. When
a new utterance \(x_{t+1}\) arrives, our goal is to update \(T_t\) to
\(T_{t+1}\) while preserving the Segment Tree properties.

\paragraph{Rightmost frontier.}
Let the rightmost frontier of \(T_t\) be denoted by \(F_t\), defined as the
ordered collection of nodes whose associated intervals have right endpoint equal
to \(t\), i.e.,
\begin{align}
F_t
=
\operatorname{\mathsf{Frontier}}(T_t)
=
(f_1,\ldots,f_m),
\qquad
\{f_1,\ldots,f_m\}
=
\{v\in V_t:r(v)=t\}.
\end{align}
The frontier nodes are ordered from lower to higher levels: \(f_1\) is the
rightmost leaf and \(f_m\) is the root. Under the levelled-tree invariant,
\(F_t\) contains exactly one active node per level. Since \(x_{t+1}\) has
interval \([t+1,t+1]\), it can only attach to a node whose interval ends at
\(t\). Thus, the admissible insertion candidates are precisely the non-leaf nodes in
\(F_t\), i.e., \(F_t\setminus\{f_1\}\), giving the update local scope rather than
requiring a scan over all tree nodes.

\begin{wrapfigure}{r}{0.5\columnwidth}
  \vspace{-1.2em}
  \centering
\includegraphics[width=0.5\textwidth]{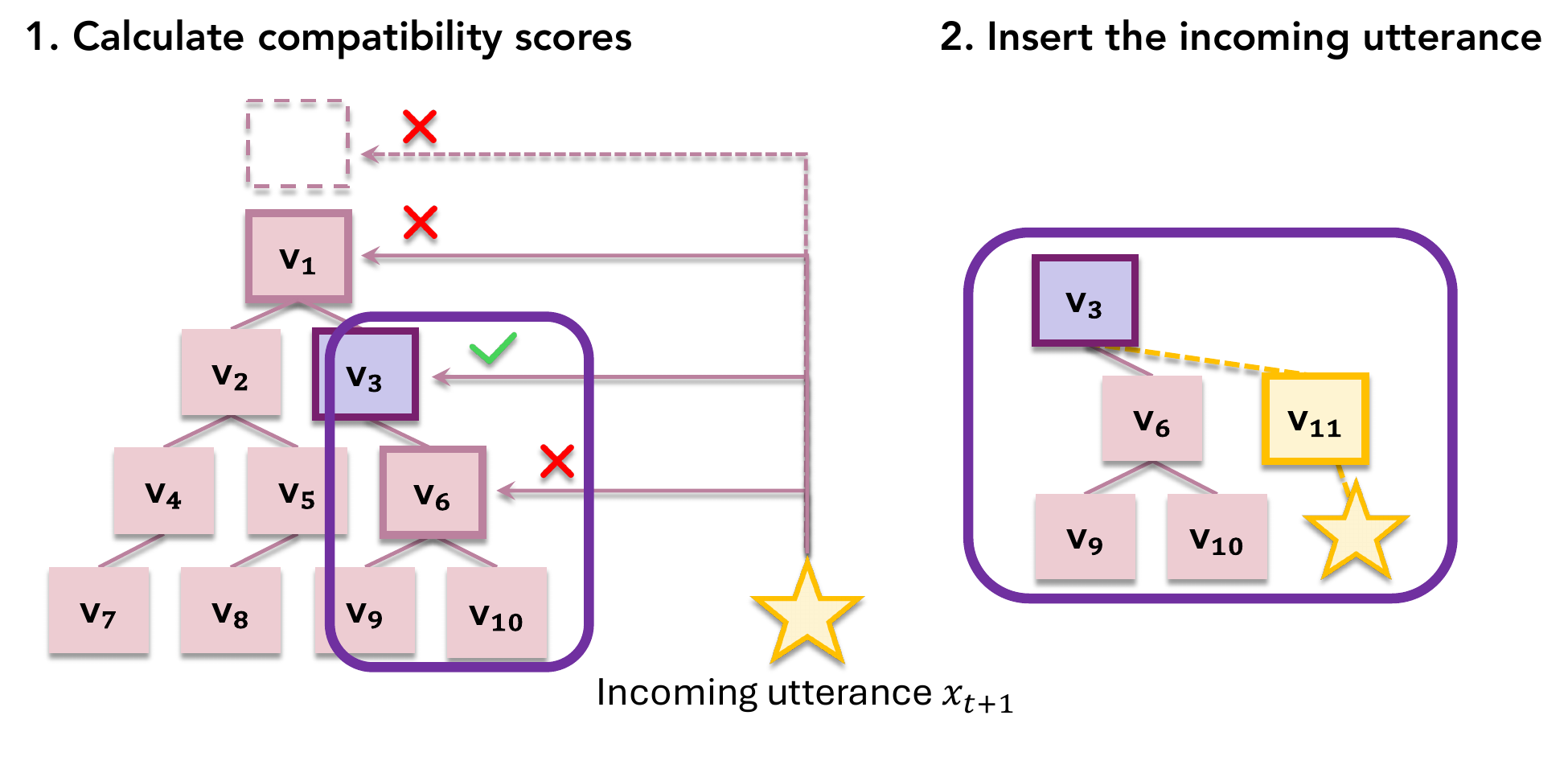}
  \vspace{-1.5em}
\caption{Online Segment Tree update. For a new utterance \(x_{t+1}\), the compatibility model selects a frontier node \(v_3\), to which a subtree with leaf \(x_{t+1}\) is attached.}
  \label{fig:online_updates}
\end{wrapfigure}

\paragraph{Online tree update via rightmost frontier.}
To select an attachment node, we use a compatibility model, which either
returns the non-leaf frontier node most compatible with the incoming utterance
\(x_{t+1}\), or indicates that no compatible frontier node exists. If a frontier
node is selected, let \(f_{\ell^*}\), where \(\ell^*\in\{2,\ldots,m\}\), denote
the selected node. We create a chain of \(\ell^*-2\) internal nodes above the new
leaf \(z_{t+1}\), each with interval \([t+1,t+1]\), and attach the chain top, or
\(z_{t+1}\) itself when \(\ell^*=2\), as the rightmost child of
\(f_{\ell^*}\). We then update the right endpoints and annotations of
\(f_{\ell^*}\) and its ancestors. If no frontier node is compatible, we create a
new root \(f_{m+1}\) whose leftmost child is the previous root of \(T_t\). We
also create a chain of \(m-1\) internal nodes above \(z_{t+1}\), each with
interval \([t+1,t+1]\), and attach the chain top, or \(z_{t+1}\) itself when
\(m=1\), to \(f_{m+1}\). The interval of \(f_{m+1}\) is then updated to
\([1,t+1]\), and its annotation is created accordingly. The algorithm is
visualized in \Autoref{fig:online_updates}.


\paragraph{Design choices for memory construction.}
The design choices for online memory construction center on the compatibility
model, which determines where a new utterance is inserted. We consider three
forms of compatibility. (i) \emph{\textbf{Cosine similarity}} selects the frontier node whose annotation embedding is most similar to the new utterance. (ii) \emph{\textbf{Pointwise LLM}} \citep{pointwisejudge1,pointwisejudge2,gprllm2025,bagel2026}independently scores each frontier node annotation against the new utterance and selects the most compatible node. (iii) \emph{\textbf{Batch LLM}} ~\citep{ma2023zeroshot,sun2023chatgpt,zhang2023rankwithoutgpt} presents all frontier node annotations with the new utterance to an LLM, which jointly compares candidates and selects the most compatible node.

\begin{figure*}[t]
  \centering
  \includegraphics[width=0.9\textwidth]{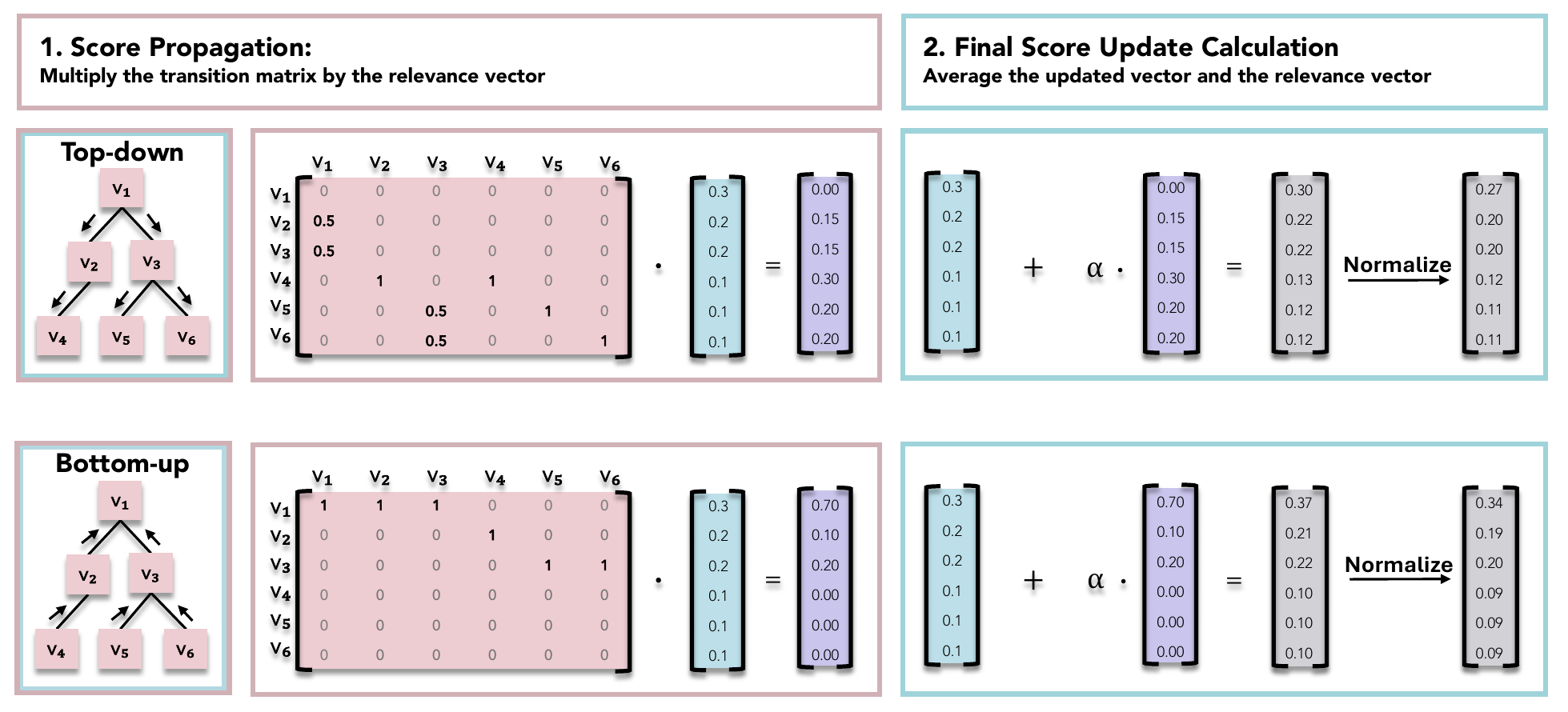}
  \caption{Two score propagation policies: top: top-down, where scores propagate
  from parents to children; bottom: bottom-up, where scores propagate from
  children to parents. The matrices illustrate the corresponding propagation
  operators. Blue arrays denote normalized initial local scores, purple arrays
  denote scores after one propagation step, and gray arrays denote the final
  scores obtained by combining the initial and propagated scores with
  \(\alpha=0.1\).}
  \label{fig:retrieval}
\end{figure*}


\subsection{Retrieval from the Memory Tree}
\label{sec:retrieval}

Our retrieval operator follows a simple principle: identify memory units relevant to the query, then expand around them to gather context. This is motivated by the fact that relevant information typically appears within coherent temporal spans rather than in isolation. Accordingly, the operator satisfies the retrieval desiderata in \Autoref{sec:preliminaries}: (i) it assigns initial scores that give higher values to more relevant nodes, selecting them as entry points; (ii) it propagates these scores over the tree, allowing high-scoring nodes to expand into their structurally and temporally admissible neighborhoods.

\subsubsection{Structure-Aware Score Propagation}

To incorporate the hierarchical structure of the memory tree, we propagate
relevance scores through structurally meaningful connections using a
matrix-based multi-hop formulation, inspired by classical graph-based ranking
methods such as PageRank~\citep{pagerank1999,PPR2003,randomwalk2006}. 

Formally, a propagation policy \(P\) induces a sparse propagation matrix
\(W_P \in [0,1]^{|V|\times |V|}\), where each entry
\(
[W_P]_{uv} = \Pr(u \mid v, q)
\)
gives the probability of propagating relevance from node \(v\) to node \(u\)
given query \(q\). Thus, \([W_P]_{uv}\) captures the contribution
of node \(v\) to node \(u\) in a single propagation step, while repeated
applications of \(W_P\) enable multi-hop aggregation over the tree structure.

Two canonical propagation policies are shown in \Autoref{fig:retrieval}: (i) \emph{top-down}, which distributes relevance from a
node to its children, and (ii) \emph{bottom-up}, which propagates relevance
from a node to its parent.

\paragraph{Initial score normalization.}
Given a query \(q\), retrieval aims to identify nodes whose associated
conversation spans are relevant for answering \(q\). Each node \(v \in V\) is
represented by its annotation \(A(v)\), from which we compute a local relevance
score $r_q(v) = R(q, A(v)) \in \mathbb{R}.$
We normalize these scores to obtain an initial query distribution as
$
s_q^{(0)}(v) = \frac{r_q(v)}{\sum_{u \in V} r_q(u)}.$

This normalization ensures that \(s_q^{(0)}\) forms a valid probability
distribution, which stabilizes subsequent matrix-based propagation and prevents
scale explosion or vanishing across multiple hops. 
Selecting nodes directly based on \(s_q^{(0)}\) recovers
\emph{collapsed-tree} retrieval~\citep{memtree2024,raptor2024}, which ignores
the tree structure and treats nodes independently.

\paragraph{Score propagation.}
Starting from the initial distribution \(s_q^{(0)} \in \mathbb{R}^{|V|}\), we define the \(k\)-hop
structural distribution via iterative propagation as
\begin{align}
s_q^{(k)} = W_P s_q^{(k-1)} = W_P^k s_q^{(0)},
\qquad k = 1, \ldots, H.
\end{align}
Each multiplication by \(W_P\) corresponds to one structural transition along
the tree.

\paragraph{Final retrieval score.}
We define the final structure-aware retrieval score as a finite-horizon
combination of the local and propagated distributions:
\begin{align}
\tilde{s}_q
=
\frac{\sum_{k=0}^{H} \alpha^k s_q^{(k)}}
{\sum_{k=0}^{H} \alpha^k},
\qquad 
0 \le \alpha < 1,
\label{eq:structure_aware_score}
\end{align}

where \(\alpha\) controls the relative contribution of longer propagation
paths. The resulting vector \(\tilde{s}_q \in \mathbb{R}^{|V|}\) assigns a structure-aware relevance score to each node. The retriever then selects
the top-\(K\) nodes with the largest values in  \(\tilde{s}_q \in \mathbb{R}^{|V|}\) .


\paragraph{Design choices for retrieval.}
We consider the following design choices for the retrieval mechanism:
the relevance scorer \(R(\cdot, \cdot)\), the propagation policy
\(P \in \{\mathrm{no\text{-}propagation}, \mathrm{top\text{-}down}, \mathrm{bottom\text{-}up}\}\),
the induced transition matrix \(W_P\), the propagation horizon \(H\), and the
decay factor \(\alpha\).

\section{Experiments}
\label{sec:experiments}

\subsection{Research Questions}
\label{sec:research_questions}



\textbf{RQ1: Does temporal tree construction improve final performance?}
We compare Segment Tree construction with non-temporal similarity clustering
trees under a temporal-order permutation experiment with matched retrieval settings to evaluate whether preserving temporal
order in the constructed memory tree improves final performance.

\textbf{RQ2: Which online construction strategy gives the best performance-cost
tradeoff?}
We compare cosine similarity, pointwise LLM, and batch LLM as compatibility
models for online Segment Tree construction, and evaluate their final
performance together with tree construction latency and cost.

\textbf{RQ3: How do tree-aware retrieval policies affect retrieval quality?}
We compare no-propagation, top-down, and bottom-up retrieval policies, and study
how propagation horizon \(H\) and decay factor \(\alpha\) contribute to each
policy's final performance.

\textbf{RQ4: How does full \ours\ perform against existing memory baselines?}
We compare the full \ours\ system against flat retrieval, tree-based memory, and
structured memory baselines to evaluate its end-to-end conversational memory
performance.

\subsection{Datasets and Evaluation}
\label{sec:datasets}

We evaluate on three long-horizon memory benchmarks: \textbf{LoCoMo} with
10 conversation groups and 1,986 queries~\citep{locomo2024},
\textbf{LongMemEval-MAB} with 5 context groups and 300 queries
~\citep{longmemeval2024,memoryagentbench2025}, and \textbf{RealMem} with
10 personas and 1,415 queries~\citep{realmem2026}. LongMemEval-MAB denotes the
MemoryAgentBench reformulation of LongMemEval(S), where memory is built over an
extended dialogue and queried repeatedly. We use \texttt{text-embedding-3-small} for all
embedding-based retrieval and similarity computations. We report LLM-judge
accuracy and token-level F1. For LLM-dependent components, including LLM-based
memory construction and answer generation, we evaluate two backbones:
\texttt{gpt-5.4-mini} \citep{gpt54mini2026} and \texttt{qwen-3.5-flash} \citep{qwen35flash2026}. Answer correctness is judged by \texttt{gpt-4o-mini} \citep{gpt4omini2024}. All methods use the same evidence budget \(K=10\). We only report \texttt{gpt-5.4-mini} results in the main paper. Full \texttt{qwen-3.5-flash} results are provided in \appref{sec:appendix-qwen-main-results} and category-level breakdowns are provided in \appref{sec:appendix-question-type-breakdown}.

\subsection{Baselines and Configurations}
\label{sec:baselines}

We compare against three groups of external baselines. \textbf{Flat retrieval}
baselines include BM25 and Dense retrieval~\citep{bm252009,textembedding3small2024}.
\textbf{Tree-based} baselines include RAPTOR and MemTree
~\citep{raptor2024,memtree2024}. \textbf{Structured memory} baselines include
A-MEM, Mem0, and HippoRAG~\citep{amem2025,mem02025,hipporag2024}. We do not
include MemWalker and LATTICE as direct baselines because they do not provide
readily reproducible implementations for our online conversational setting
~\citep{memwalker2023,lattice2025}. See \appref{sec:implementation_details} for full implementation details.

\paragraph{\ours\ variants.}

\begin{wraptable}{r}{0.47\textwidth}
\vspace{-1.2em}
\centering
\scriptsize
\setlength{\tabcolsep}{3pt}
\renewcommand{\arraystretch}{1.08}
\begin{tabular}{@{}p{0.38\linewidth}p{0.56\linewidth}@{}}
\toprule
\rowcolor{lightpurple}
\textbf{Design choice} & \textbf{Value space} \\
\midrule

\rowcolor{lightpurple!45}
\multicolumn{2}{@{}l}{\textbf{Online construction}} \\
Compatibility model \(C\) &
cosine similarity, pointwise LLM, batch LLM \\

\addlinespace[1pt]
\rowcolor{lightpurple!45}
\multicolumn{2}{@{}l}{\textbf{Retrieval}} \\
Local relevance \(R\) &
normalized cosine similarity \\
Propagation policy \(P\) &
no-propagation, top-down, bottom-up \\
Transition matrix \(W_P\) &
binary structural weights, column-normalized \\
Decay factor \(\alpha\) &
\(\{.05,.10,.20,.30,.50,.70,.95\}\) \\
Propagation horizon \(H\) &
\(\{0,1,2,3\}\) \\
\bottomrule
\end{tabular}
\vspace{-0.4em}
\caption{\small Experimental setting of \ours.}
\label{tab:ours_config}
\vspace{-1.2em}
\end{wraptable}
\footnote{Anonymized implementation and reproduction instructions are available at
\url{https://anonymous.4open.science/r/SegTreeMem-B325/README.md}.}
We evaluate \ours\ by varying online construction and tree-aware retrieval
settings, as summarized in \Autoref{tab:ours_config}. For online construction,
we compare three compatibility models: cosine similarity, pointwise LLM, and
batch LLM. For retrieval, we compute node-query relevance with
\texttt{text-embedding-3-small}, normalize the local scores into an initial query
distribution, and compare three propagation policies \(P\): no-propagation,
top-down, and bottom-up. In our experiments, \(W_P\) is set by assigning binary
raw weights to the allowed source-target transitions induced by \(P\)
(\Autoref{fig:retrieval}), and then column-normalizing these weights.
\subsection{Results}
\label{sec:results}

\textbf{RQ1: Does temporal tree construction improve final performance?}
\begin{figure*}[t]
  \centering
  \includegraphics[width=0.9\textwidth]{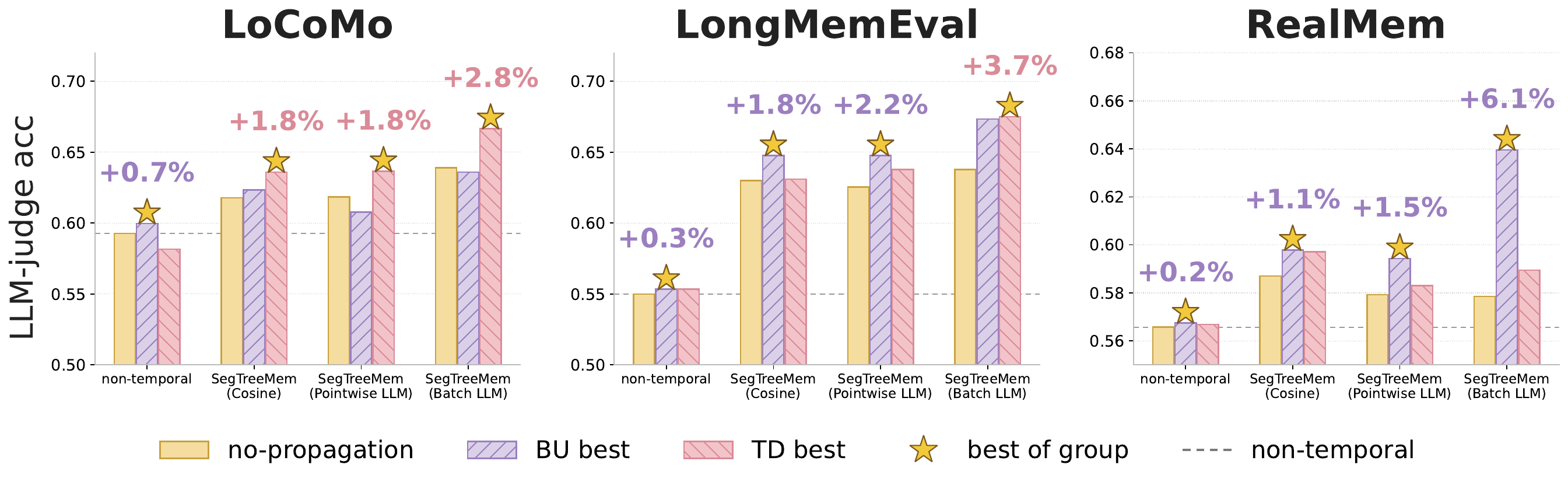}
    \caption{
    Controlled comparison of tree construction strategies. We compare a
    non-temporal similarity clustering tree with Segment Tree variants under
    matched retrieval settings. BU and TD report the best bottom-up and top-down
    configurations. Segment Tree construction consistently improves over the
    non-temporal baseline, showing the benefit of preserving temporal order for memory construction.
    }
  \label{fig:treebuild_control}
\end{figure*}

\paragraph{Temporal trees improve final performance over non-temporal baselines.}
In \Autoref{fig:treebuild_control}, we compare Segment Tree with a
non-temporal similarity clustering tree. See
\appref{sec:appendix-nontemporal-baseline} for the implementation of this
baseline. We control the node-level summarization prompt and other implementation
details, so the only difference is the online update algorithm. Across the three datasets, Segment Tree variants consistently outperform
the non-temporal tree, and applying the same propagation policies to the
non-temporal tree does not close the gap. This suggests that preserving temporal
order during construction yields a stronger memory representation for
conversational retrieval.

\paragraph{Temporal tree construction improves internal-node coherence.}
To investigate why temporal tree construction improves over the non-temporal
similarity clustering tree, we conduct a qualitative analysis of tree instances (\appref{sec:appendix-tree-analysis}). We find that Segment Tree internal nodes tend to be more
semantically coherent and fluent because they summarize temporally ordered
utterance spans. In contrast, similarity clustering can group utterances from
distant parts of a conversation under the same internal node, which may lead to
less coherent summaries. 

\begin{wraptable}{r}{0.52\columnwidth}
\vspace{-1.0em}
\centering
\scriptsize
\setlength{\tabcolsep}{2.4pt}
\renewcommand{\arraystretch}{1.05}
\begin{tabular}{llccc}
\toprule
\rowcolor{lightpurple}
\textbf{Tree} & \textbf{Order}
& \textbf{LoCoMo}
& \textbf{LME}
& \textbf{RealMem} \\
\midrule
\multirow{3}{*}{Non-temp.}
& Orig. & 0.603 & 0.595 & 0.574 \\
& Perm. & 0.583 & 0.576 & 0.523 \\
& \(\Delta\) & -0.020 & -0.019 & -0.051 \\
\midrule
\multirow{3}{*}{\ours{}}
& Orig. & 0.639 & 0.630 & 0.580 \\
& Perm. & 0.538 & 0.523 & 0.436 \\
& \(\Delta\) & -0.111 & -0.107 & -0.144 \\
\bottomrule
\end{tabular}
\caption{Temporal-order permutation experiment under batch-LLM construction and no-propagation retrieval. Scores are LLM-judge accuracy.}
\label{tab:temporal_permutation}
\vspace{-1.0em}
\end{wraptable}
\paragraph{Temporal permutation confirms the contribution of temporal order.}
To test whether temporal order is actually used by the constructed memory, we conduct a temporal-order permutation experiment by swapping 30\% of randomly selected turn pairs before memory construction and rebuilding each tree from the permuted sequence. We use batch-LLM construction with no-propagation retrieval. As shown in \Autoref{tab:temporal_permutation}, permutation only mildly affects
the non-temporal tree but causes a much larger performance drop for \ours{}. This indicates that preserving temporal order during construction contributes directly to \ours{}'s performance.


\textbf{RQ2: Which online construction strategy gives the best performance-cost
tradeoff?}

\paragraph{Batch LLM gives the best construction quality.}
RQ2 compares the compatibility models used during online Segment Tree
construction. As shown in \Autoref{fig:treebuild_control}, batch LLM is consistently strong under no-propagation and also receives larger gains from tree-aware retrieval. This suggests that batch LLM produces
coherent memory nodes that perform well without propagation and yields a
tree structure that can be better exploited by propagation. This is consistent
with findings in LLM-based ranking that listwise comparison can better support
relative judgments among candidates than independent pointwise scoring
\citep{ma2023zeroshot,sun2023chatgpt}.

\begin{figure*}[t]
  \centering
  \includegraphics[width=0.98\textwidth]{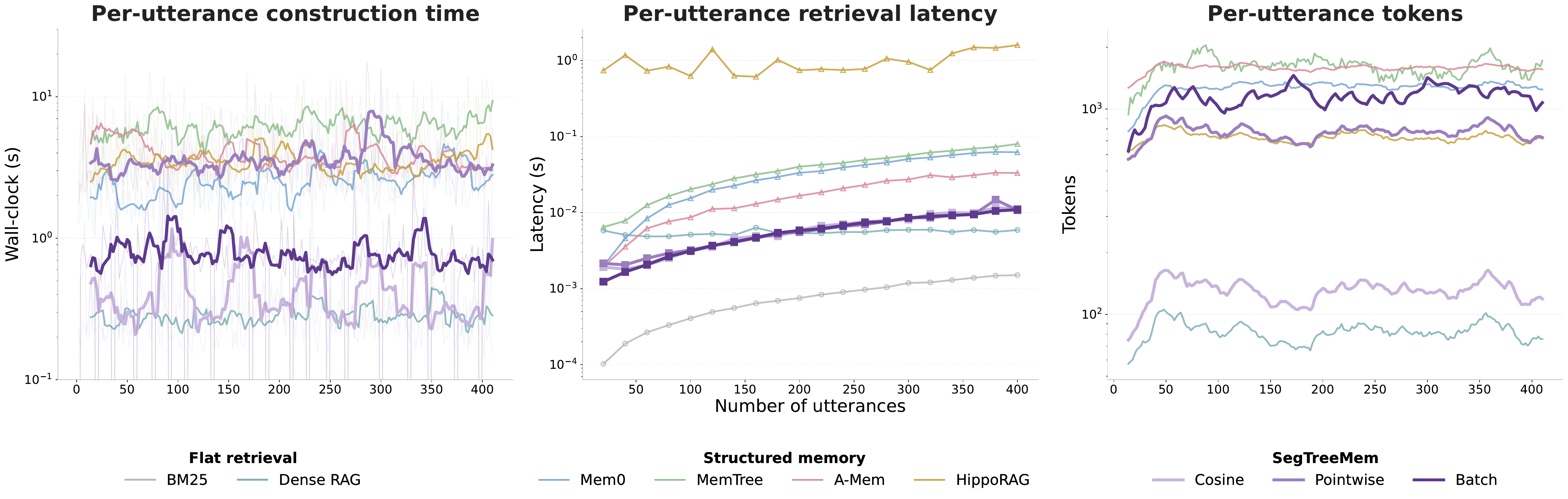}
    \caption{Construction and retrieval efficiency as memory grows. We report
    per-utterance construction time, per-utterance retrieval latency, and
    per-utterance token usage for flat retrieval baselines, structured memory
    baselines, and \ours\ construction variants. \ours{} remains efficient among
    structured memory methods, with cosine construction providing the lowest-cost
    variant and LLM-based construction remaining competitive in runtime and token
    usage.}
  \label{fig:runtime}
\end{figure*}

\paragraph{Cosine provides the better quality-cost tradeoff.}
Construction quality should be interpreted together with online cost
(\Autoref{fig:runtime}). Batch LLM gives the strongest quality, but it requires
LLM-based frontier scoring. In contrast, cosine similarity uses no LLM calls for frontier
matching, and the runtime analysis shows roughly \(10\times\) lower token usage
and about \(5\times\) lower construction latency than batch LLM in the online
construction trace. Cosine similarity therefore provides the better quality-cost trade-off as it remains competitive in accuracy while better satisfying the real-time update requirement. Thus, batch LLM is preferable when construction quality is
prioritized, while cosine similarity is the practical choice when latency or token budget
is the main constraint.


\textbf{RQ3: How do tree-aware retrieval policies affect retrieval quality?}

\begin{figure*}[t]
  \centering
  \includegraphics[width=0.98\textwidth]{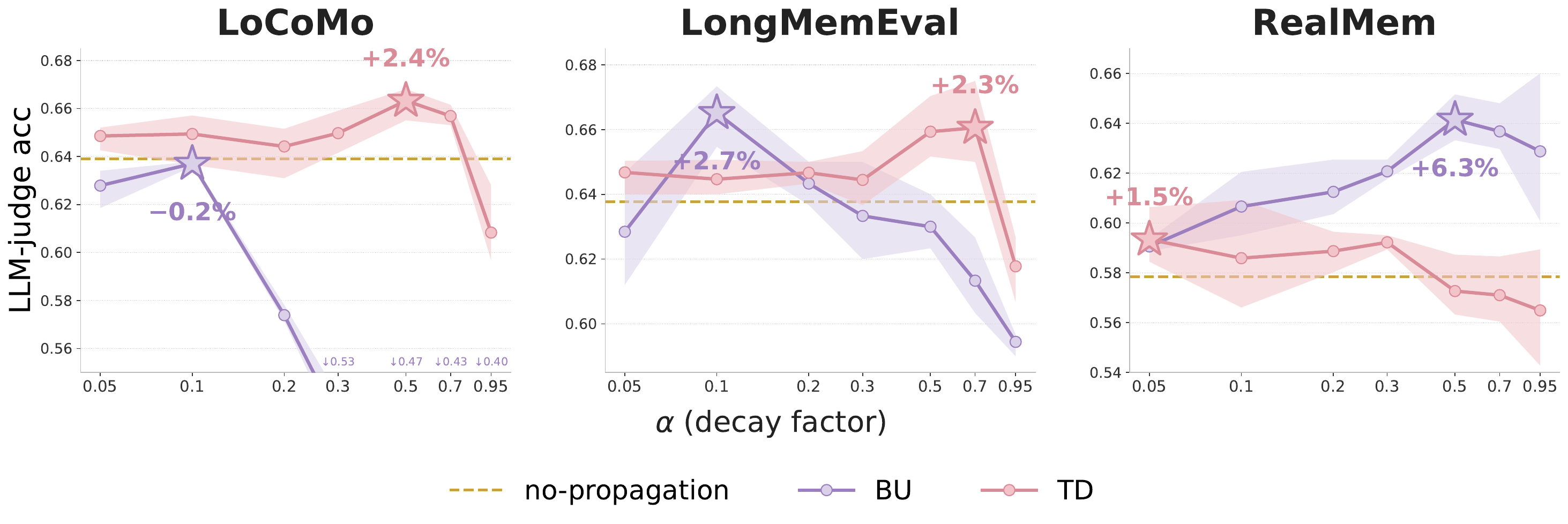}
    \caption{
    Effect of propagation policy and decay factor on retrieval quality. The dashed
    line denotes no-propagation retrieval, corresponding to \(H=0\). Curves show
    bottom-up and top-down propagation as \(\alpha\) varies, with performance
    averaged over \(H\in\{1,2,3\}\). All results use batch LLM tree construction.
    Moderate propagation often improves retrieval, while the stronger direction
    depends on the temporal granularity required by each dataset.
}
  \label{fig:propagation_sweep}
\end{figure*}

\begin{figure*}[t]
    \centering
    \includegraphics[width=\textwidth]{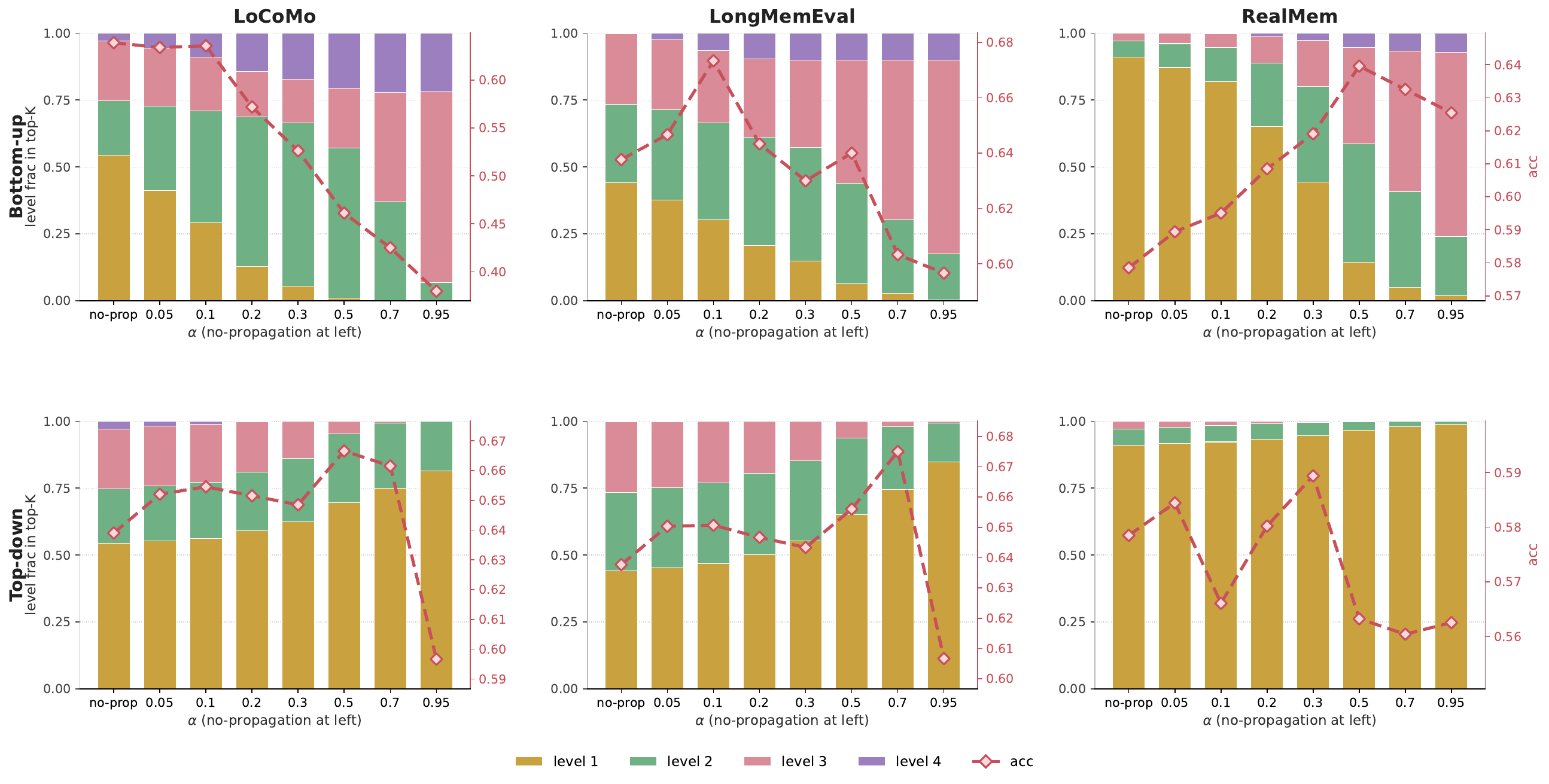}
    \caption{Retrieved-node level distribution as a function of the propagation
    setting, using batch-LLM \ours{} trees and a fixed propagation horizon
    \(H=2\). Stacked bars show the fraction of retrieved nodes at each dataset-specific temporal level, and the dashed line shows the corresponding accuracy. Bottom-up propagation increasingly shifts retrieval toward broader
    internal nodes as \(\alpha\) grows, whereas top-down propagation remains
    more concentrated on exchange-level evidence.}
    \label{fig:appendix-retrieval-levels}
\end{figure*}

\paragraph{Retrieval behavior is dataset-dependent.}
\Autoref{fig:propagation_sweep} shows that the effect of tree-aware retrieval
depends on the dataset, propagation policy $P$, horizon \(H\), and decay factor
\(\alpha\). No single propagation configuration dominates across all settings.
Top-down propagation gives the strongest results on LoCoMo and LongMemEval-MAB,
while RealMem shows stronger gains from bottom-up propagation.

\paragraph{Propagation direction changes temporal granularity.}
To explain the dataset-specific effects, we analyze how different propagation
policies change the distribution of retrieved node levels in
\appref{sec:appendix-additional-propagation}. The analysis shows that
different propagation policies shift the temporal granularity of the retrieved
evidence. Top-down retrieval selects more specific spans, while bottom-up
retrieval increases the selection of internal nodes that cover broader temporal
contexts. This helps explain the dataset-specific behavior in
\Autoref{fig:propagation_sweep}. LoCoMo contains many detail-oriented questions,
where retrieving specific utterance spans is more useful. RealMem is more project-oriented, where
answers often depend on the broader state of an evolving interaction. In real
applications, this suggests that the propagation policy should be chosen
according to the temporal granularity required by the task.

\paragraph{Decay factor \(\alpha\) and horizon \(H\) tune probability allocation.}
\Autoref{fig:propagation_sweep} shows that the best propagation setting differs
across datasets and policies. Moderate propagation often improves over
no-propagation, while very large \(\alpha\) can sharply reduce accuracy. To
better understand the effects of \(\alpha\) and \(H\), we provide full
\(H\)-sweep results and analyze retrieved node levels under different
propagation policies and \((\alpha,H)\) settings in
\appref{sec:appendix-additional-propagation}. The analysis shows that
\(\alpha\) and \(H\) determine how far retrieval moves from the initial topical
entry distribution and how strongly the retrieved evidence follows the
corresponding policy direction. Larger \(\alpha\) and larger \(H\) push
retrieval further along that direction. Thus, the best \((\alpha,H)\) setting
depends on the temporal granularity required by the task and how far retrieval
should move from the initial distribution.


\textbf{RQ4: How does full \ours\ perform against existing memory baselines?}


\definecolor{lightpurple}{RGB}{244,238,255}
\definecolor{headerpurple}{RGB}{226,214,245}
\definecolor{cigray}{RGB}{95,95,95}
\definecolor{groupgray}{RGB}{80,80,80}

\newcommand{\scoreci}[2]{#1 {\scriptsize\textcolor{cigray}{[#2]}}}
\newcommand{\bestscoreci}[2]{\textbf{#1} {\scriptsize\textcolor{cigray}{[#2]}}}
\newcommand{\methodgroup}[1]{%
  \multicolumn{7}{@{}l}{\textit{\textcolor{groupgray}{#1}}}%
}

\begin{table*}[t]
\centering
\small
\setlength{\tabcolsep}{2.8pt}
\renewcommand{\arraystretch}{1.10}
\resizebox{\textwidth}{!}{%
\begin{tabular}{@{}lcccccc@{}}
\toprule
\rowcolor{headerpurple}
\textbf{Method} &
\multicolumn{2}{c}{\textbf{LoCoMo}} &
\multicolumn{2}{c}{\textbf{LongMemEval-MAB}} &
\multicolumn{2}{c}{\textbf{RealMem}} \\
\rowcolor{headerpurple}
&
\textbf{LLM} & \textbf{F1} &
\textbf{LLM} & \textbf{F1} &
\textbf{LLM} & \textbf{F1} \\
\midrule

\methodgroup{Retrieval-only memory baselines} \\

BM25 \citep{bm251994}
& \scoreci{0.469}{0.447, 0.491}
& \scoreci{0.184}{0.177, 0.192}
& \scoreci{0.497}{0.440, 0.553}
& \scoreci{0.169}{0.150, 0.189}
& \scoreci{0.498}{0.472, 0.524}
& \scoreci{0.317}{0.312, 0.322} \\

Dense \citep{dense2020}
& \scoreci{0.490}{0.468, 0.512}
& \scoreci{0.188}{0.180, 0.196}
& \scoreci{0.520}{0.463, 0.577}
& \scoreci{0.168}{0.149, 0.187}
& \scoreci{0.524}{0.498, 0.550}
& \scoreci{0.316}{0.311, 0.321} \\

\midrule

\methodgroup{Graph-structured memory baselines} \\

A-MEM \citep{amem2025}
& \scoreci{0.608}{0.586, 0.629}
& \scoreci{0.231}{0.223, 0.239}
& \scoreci{0.553}{0.497, 0.610}
& \scoreci{0.166}{0.148, 0.184}
& \scoreci{0.515}{0.488, 0.541}
& \scoreci{0.308}{0.303, 0.314} \\

Mem0 \citep{mem02025}
& \scoreci{0.454}{0.432, 0.476}
& \scoreci{0.187}{0.179, 0.194}
& \scoreci{0.537}{0.480, 0.593}
& \scoreci{0.174}{0.156, 0.191}
& \scoreci{0.450}{0.424, 0.476}
& \scoreci{0.314}{0.310, 0.318} \\

HippoRAG \citep{hipporag2024}
& \scoreci{0.570}{0.548, 0.592}
& \scoreci{0.218}{0.210, 0.226}
& \scoreci{0.560}{0.504, 0.616}
& \scoreci{0.172}{0.153, 0.190}
& \scoreci{0.531}{0.505, 0.557}
& \scoreci{0.310}{0.304, 0.315} \\

\midrule

\methodgroup{Tree-structured memory baselines} \\

RAPTOR \citep{raptor2024}
& \scoreci{0.593}{0.571, 0.614}
& \scoreci{0.223}{0.213, 0.232}
& \scoreci{0.523}{0.467, 0.580}
& \scoreci{0.166}{0.148, 0.185}
& \scoreci{0.579}{0.553, 0.604}
& \scoreci{0.312}{0.306, 0.318} \\

MemTree \citep{memtree2024}
& \scoreci{0.483}{0.461, 0.505}
& \scoreci{0.151}{0.144, 0.157}
& \scoreci{0.483}{0.427, 0.540}
& \scoreci{0.157}{0.139, 0.175}
& \scoreci{0.536}{0.510, 0.562}
& \scoreci{0.310}{0.304, 0.315} \\

\midrule
\methodgroup{Proposed method} \\
\rowcolor{lightpurple}
\textbf{\ours}-TD
& \scoreci{0.636}{0.615, 0.657}
& \bestscoreci{0.308}{0.302, 0.314}
& \scoreci{0.633}{0.577, 0.686}
& \bestscoreci{0.204}{0.188, 0.220}
& \scoreci{0.608}{0.583, 0.634}
& \bestscoreci{0.356}{0.348, 0.363} \\
  
\rowcolor{lightpurple}
\textbf{\ours}-BU
& \bestscoreci{0.639}{0.618, 0.660}
& \scoreci{0.301}{0.295, 0.307}
& \bestscoreci{0.647}{0.591, 0.699}
& \scoreci{0.197}{0.181, 0.213}
& \bestscoreci{0.621}{0.595, 0.645}
& \bestscoreci{0.356}{0.348, 0.363} \\

\bottomrule
\end{tabular}%
}
\captionsetup{skip=10pt}
\caption{Main answer-quality results across memory benchmarks. Each reported
score includes its 95\% confidence interval in brackets. For \ours{}, TD and BU
use the same fixed representative setting (batch LLM construction,
\(\alpha=0.10\), \(H=2\)), with top-down and bottom-up propagation respectively.
\ours{} achieves the strongest overall performance across benchmarks, showing
that temporally ordered tree memory improves long-horizon agentic memory.}
\label{tab:main_results}
\end{table*}

\Autoref{tab:main_results} shows that the full \ours\ system achieves the best
LLM-judge accuracy across the evaluated benchmarks, improving over the strongest
external baseline nearly \(20\%\). The comparison supports the central claim
of this work: long-horizon conversational agents benefit from memory indexes
that are both temporal and hierarchical. Online Segment Tree construction
creates temporally ordered memory units, while tree-aware retrieval uses this
structure to recover relevant context at the appropriate granularity. \Autoref{fig:runtime} shows that  
\ours\ is also computationally competitive. Per-utterance     
  construction and retrieval is the fastest among methods that build any      
  structured memory. Token cost ranges from negligible for      
  Cosine to comparable 
  with existing structured baselines for the Pointwise 
  and Batch variants.


\section{Conclusion}
\label{sec:conclusion}

We introduced \ours, a memory architecture that organizes long-horizon
conversation history as an online Segment Tree over utterances and retrieves
evidence through tree-aware propagation. The central takeaway is that
conversational agents benefit from memory indexes that are both temporal and
hierarchical: temporal order helps construct coherent memory states, while the
tree structure lets retrieval select context at the appropriate granularity.
Our results support temporal and hierarchical indexing as a practical foundation
for long-horizon conversational memory.
\paragraph{Limitation and future work.}
Our current implementation leaves several practical extensions for future work.
First, propagation uses fixed policies and horizons across queries, although
different questions may require different retrieval granularity. Query-dependent
policy selection could choose among no propagation, top-down retrieval, and
bottom-up retrieval. Second, the current system focuses on text-based
conversational memory. Extending the same temporal ordering principle to
multimodal observations, tool calls, and action traces is an important direction
for long-horizon agents. Third, future systems could add incremental maintenance
operations, such as local subtree rebuilding, periodic compression, or duplicate
cleanup, to keep the memory index efficient and stable over very long
deployments.
\bibliographystyle{plainnat}
\bibliography{references}
\clearpage
\appendix

\section{Algorithmic Details}
\label{sec:appendix-algorithmic-details}

{\raggedbottom
This subsection provides pseudocode for the two \ours{} operations. Node annotations are denoted by \(A(v)\), node intervals by \(I(v)\), and the admissible retrieval set by \(\mathcal G\).

\begin{algorithm}
\caption{Online Segment Tree Update}
\label{alg:online_tree_update}
\begin{algorithmic}[1]
\STATE \textbf{Input:} current tree $T_t=(V_t,E_t,\rho_t)$, incoming utterance $x_{t+1}$, compatibility model $C$, annotation model $S$
\STATE \textbf{Output:} updated tree $T_{t+1}$
\STATE Create a new leaf $z_{t+1}$ with $I(z_{t+1})=[t+1,t+1]$ and $A(z_{t+1})=x_{t+1}$
\IF{$T_t=\emptyset$}
    \STATE \textbf{return} the single-node tree rooted at $z_{t+1}$
\ENDIF
\STATE $F_t=(f_1,\ldots,f_m)\leftarrow\operatorname{\mathsf{Frontier}}(T_t)$, ordered from leaf to root.
\STATE $\ell^\star\leftarrow C(x_{t+1},F_t\setminus\{f_1\})$, where $\ell^\star\in\{2,\ldots,m\}\cup\{\texttt{NONE}\}$
\IF{$\ell^\star\neq\texttt{NONE}$}
    \STATE Create a rightmost chain $Z$ of $\ell^\star-2$ internal nodes above $z_{t+1}$, each with interval $[t+1,t+1]$
    \STATE Attach the chain top (or $z_{t+1}$ itself, if $\ell^\star=2$) as the rightmost child of $f_{\ell^\star}$
    \STATE $U\leftarrow Z\cup\{f_{\ell^\star}\}\cup\operatorname{\mathsf{anc}}(f_{\ell^\star})$
\ELSE
    \STATE Create a new root $\rho_{t+1}$ whose leftmost child is $\rho_t$
    \STATE Create a rightmost chain $Z$ of $m-1$ internal nodes above $z_{t+1}$, each with interval $[t+1,t+1]$
    \STATE Attach the chain top (or $z_{t+1}$ itself, if $m=1$) as the rightmost child of $\rho_{t+1}$
    \STATE $U\leftarrow Z\cup\{\rho_{t+1}\}$
\ENDIF
\FOR{each node $u\in U$ in bottom-up order}
    \STATE $I(u)\leftarrow\bigl[\min_{c\in\mathsf{ch}(u)}l(c),\,\max_{c\in\mathsf{ch}(u)}r(c)\bigr]$
    \STATE $A(u)\leftarrow S\bigl((A(c))_{c\in\mathsf{ch}(u)}\bigr)$
\ENDFOR
\STATE \textbf{return} $T_{t+1}$
\end{algorithmic}
\end{algorithm}

\paragraph{Insertion invariant.}
The update in \Autoref{alg:online_tree_update} never reorders existing leaves. The incoming utterance is always introduced as the new rightmost leaf, and only nodes on the affected right frontier are modified. Therefore every internal node continues to summarize a contiguous interval, while the update scope remains
localized to the selected frontier segment and its ancestors. The
\texttt{NONE} branch is used when the compatibility model finds no coherent
continuation among the active frontier nodes. In that case the previous tree and
the new rightmost chain are combined under a new root.

\begin{algorithm}[H]
\caption{Structure-Aware Retrieval}
\label{alg:structure_aware_retrieval}
\begin{algorithmic}[1]
\STATE \textbf{Input:} memory tree $T=(V,E)$, query $q$, admissible set $\mathcal G\subseteq V$, local relevance model $R$, propagation matrix $W_P$, horizon $H$, decay $\alpha\in[0,1)$, evidence budget $K$
\STATE \textbf{Output:} retrieved node set $\widehat{\mathcal V}_K$
\FOR{each node $v\in V$} 
    \STATE $r_q(v)\leftarrow R(q,A(v))$ \hfill \texttt{//Relevance score computation}
\ENDFOR
\STATE $s_q^{(0)}(v)\leftarrow r_q(v)\big/\sum_{u\in V}r_q(u)$ for all $v\in V$ \hfill \texttt{//Initial score normalization}
\STATE $\tilde{s}_q\leftarrow s_q^{(0)}$
\STATE $s_q^{\mathrm{prev}}\leftarrow s_q^{(0)}$
\FOR{$k=1,\ldots,H$}  
    \STATE $s_q^{(k)} \leftarrow W_P\: s_q^{\mathrm{prev}}$ \hfill \texttt{// Score propagation}
    \STATE $\tilde{s}_q \leftarrow \tilde{s}_q+\alpha^k s_q^{(k)}$ \hfill \texttt{// Propagated score accumulation}
    \STATE $s_q^{\mathrm{prev}}\leftarrow s_q^{(k)}$
\ENDFOR
\STATE $\widehat{\mathcal V}_K\leftarrow\operatorname{TopK}_{v\in\mathcal G}\tilde{s}_q(v)$ \hfill \texttt{// Retrieving nodes with high final scores}
\STATE \textbf{return} $\widehat{\mathcal V}_K$
\end{algorithmic}
\end{algorithm}

\paragraph{Retrieval invariant.}
\Autoref{alg:structure_aware_retrieval} first scores every node independently
using its annotation, then mixes the local score distribution with propagated
distributions obtained by applying $W_P$ for up to \(H\) steps. Setting
\(H=0\) recovers collapsed retrieval. For top-down propagation, probability mass
moves from broader summaries toward child spans. For bottom-up propagation, mass
moves from specific matches toward their enclosing temporal context. The final
Top-\(K\) selection is restricted to \(\mathcal G\), which lets the same scoring
routine support either internal-node retrieval or leaf-only evidence selection.

}
\newpage
\section{Extended Related Work}
\label{sec:appendix-extended-related-work}

\subsection{Long-Horizon Memory Benchmarks}

Long-horizon conversational memory benchmarks evaluate whether an agent can
retain, update, and retrieve information accumulated across extended
interactions. This setting differs from standard long-context RAG \citep{longbench2024,multihop2024,EQR2024, EQR22025,sceneXR2026}:
the memory state is built over a sequence of user--agent interactions, and
queries may require resolving information across temporally separated turns,
updated facts, evolving preferences, or task state.

We evaluate \ours{} on three benchmarks that directly target this setting:
\begin{itemize}
    \item \textbf{LoCoMo} evaluates very long-term conversational memory across
    multi-session dialogues, with questions that require retrieving facts,
    preferences, and events from extended interaction histories
    \citep{locomo2024}.

    \item \textbf{LongMemEval-MAB} is based on the MemoryAgentBench
    reformulation of LongMemEval, where memory is built incrementally over an
    extended dialogue and queried repeatedly throughout the interaction
    \citep{longmemeval2024,memoryagentbench2025}. This setting is especially
    relevant to online memory construction because the system must maintain a
    memory state as new utterances arrive rather than process a fixed document
    once.

    \item \textbf{RealMem} evaluates memory-driven interaction in more
    realistic user-facing scenarios, where questions may depend on evolving
    project state, user preferences, or prior task context \citep{realmem2026}.
\end{itemize}

These benchmarks are well matched to our goal because they evaluate memory as
an external state maintained over interaction histories. Other long-context
benchmarks, including synthetic needle-retrieval tasks, long-document QA, and
book-level reading comprehension, are useful for measuring whether a model can
attend over long input sequences. However, they do not directly evaluate the
central setting of this paper: online construction of a persistent
conversational memory index and retrieval from that index as the interaction
evolves. We therefore focus our experiments on benchmarks where memory
construction and retrieval are both part of the task.

\subsection{Flat Retrieval Methods}

\paragraph{Memory construction.}
Flat retrieval methods store past context as an unordered
collection of independent memory records. In sparse retrieval, records are
indexed by lexical matching signals such as BM25 \citep{bm251994}. In dense
retrieval, records are embedded into a vector space and indexed for nearest
neighbor search \citep{dense2020}. These methods are simple, scalable, and often
strong baselines for retrieval-augmented generation. However, their construction
step does not explicitly model the hierarchical or temporal structure of a
conversation. Each memory unit is usually treated as an independent item, even
when its meaning depends on surrounding utterances.

\paragraph{Retrieval.}
At retrieval time, flat methods rank memory records independently by lexical or
semantic similarity to the query. This supports topical matching but does not
explicitly use temporal or structural context. For example, if a query matches
one utterance in a longer coherent span, flat retrieval does not automatically
recover nearby utterances, enclosing summaries, or other parts of the same
interaction segment. In our terminology, flat retrieval satisfies topical
relevance but does not directly exploit the admissible temporal or structural
neighborhood around a match.

\subsection{Graph-Based and Agentic Memory Methods}

\paragraph{Memory construction.}
Graph-based and agentic memory systems enrich flat memory by extracting,
linking, updating, or consolidating memory records over time. Generative Agents
store memories in a stream and periodically synthesize higher-level reflections
\citep{generativeagents2023}.MemoryBank incorporates importance weighting and forgetting mechanisms for long-term user memory \citep{memorybank2024}. MemGPT organizes
memory across tiers and lets the agent manage memory through operations
analogous to paging \citep{memgpt2023}. More recent agentic memory systems such
as A-MEM and Mem0 emphasize explicit memory operations, including extraction,
updating, linking, and consolidation \citep{amem2025,mem02025}. Goal-oriented memory reasoning further studies how an agent can decompose a user
utterance into subgoals and retrieve memory targeted to those reasoning needs \citep{goalmem2026}.

These systems are designed for persistent agent memory, but their memory
organization is often driven by learned or heuristic estimates of memory
importance, extracted entities, or agentic update decisions rather than by a
deterministic temporal constraint.

\paragraph{Retrieval.}
Graph-structured retrieval methods use links between memory records to expand
beyond an initial topical match \citep{madpr2025,semanticxpath2026}. This idea is related to classical graph ranking
methods such as PageRank, personalized PageRank, and random walks with restart
\citep{pagerank1999,PPR2003,randomwalk2006}. HippoRAG applies this perspective
to LLM memory by constructing a corpus-derived knowledge graph and using
personalized PageRank to obtain a structure-aware relevance signal
\citep{hipporag2024}. A-MEM also uses explicit memory links to support
associative retrieval \citep{amem2025}.

Our retrieval method shares the high-level idea that relevance can be propagated
through structure. The key difference is the underlying structure. Graph-based
methods typically propagate over entity, fact, or memory-record links. In
contrast, \ours{} propagates over a temporal hierarchy whose nodes correspond to
contiguous conversational spans. This makes propagation directly interpretable
as moving between local utterances, enclosing temporal contexts, and related
neighboring spans.

\subsection{Tree-Based Memory Methods}

\paragraph{Memory construction.}
Tree-based memory methods organize text into hierarchical abstractions.
RAPTOR constructs a recursive abstraction tree by clustering text chunks and
summarizing clusters \citep{raptor2024}. LATTICE similarly builds a semantic
hierarchy for LLM-guided retrieval \citep{lattice2025}. MemWalker preserves
document order by grouping adjacent segments into fixed-size units
\citep{memwalker2023}. MemTree is designed for conversational memory and
supports online updates by inserting new memories into the most semantically
similar branch \citep{memtree2024}.

These methods differ in whether they support online construction and whether
they preserve order. RAPTOR and LATTICE build semantic abstraction trees through
clustering, which can group non-adjacent segments and therefore does not
guarantee temporal contiguity. MemWalker preserves order but is designed for
static contexts rather than online conversational updates. MemTree supports
online insertion, but its similarity-based insertion rule prioritizes topical
relatedness and does not preserve utterance order in the tree structure.
\ours{} is designed to satisfy both requirements: it updates memory online while
maintaining a tree over temporally contiguous utterance spans.

\paragraph{Retrieval.}
Existing tree-based retrieval methods generally follow one of two strategies.
RAPTOR and MemTree perform \emph{collapsed retrieval}: they flatten the tree
into a set of candidate nodes and rank all nodes independently
\citep{raptor2024,memtree2024}. This allows retrieval at multiple abstraction
levels, but it does not explicitly use the tree edges during retrieval.
MemWalker and LATTICE instead perform \emph{LLM-guided traversal}, where an LLM
moves from coarse nodes to finer-grained children to select relevant branches
\citep{memwalker2023,lattice2025}. This uses the top-down hierarchy, but it
does not directly exploit other tree-induced relations such as parent context,
sibling context, or temporal neighborhoods around an initially relevant match.

\ours{} differs from both approaches. Rather than flattening the tree or
performing a single top-down traversal, it treats retrieval as finite-horizon
relevance propagation over the memory tree. This lets the retriever begin from
topical matches and then move through structurally meaningful relations, such
as from a detailed utterance to its enclosing context or from a broad summary
to its child spans. Because every node in \ours{} corresponds to a contiguous
utterance interval, these structural moves also have a temporal interpretation.

\subsection{Conversational Segmentation and Segment Trees}

Conversational segmentation studies how to identify coherent stretches of
dialogue and topic boundaries in interaction histories \citep{latin2025}. Prior work has studied
discourse segmentation in multi-party conversation \citep{galley2003discourse}
and topic segmentation in asynchronous conversations \citep{joty2013topic}.
These methods typically produce a flat sequence of boundaries as a
preprocessing step.

The compatibility model in \ours{} performs a related decision, but in an
online and hierarchical memory construction setting. Each incoming utterance is
compared against the rightmost frontier of the current tree, where different
frontier nodes correspond to different temporal granularities. The update
therefore decides not only whether the utterance continues the current topic,
but also at what level of temporal abstraction it should be attached.

Segment trees originate as data structures for range queries over intervals
\citep{deberg2008computational}. We adapt this classical structure to
conversational memory by associating each node with an LLM-generated annotation
of a contiguous utterance span and by defining an online rightmost-frontier
update rule. To our knowledge, \ours{} is the first LLM-agent memory system to
use a Segment Tree as the primary memory index for online construction and
structure-aware retrieval over long-horizon conversations.
\newpage
\section{Tree Structure Analysis}
\label{sec:appendix-tree-analysis}

\subsection{Cross-Dataset Quantitative Statistics}
\label{sec:appendix-quant}

\begin{table}[h]
\centering
\small
\setlength{\tabcolsep}{3.8pt}
\renewcommand{\arraystretch}{1.05}
\resizebox{\linewidth}{!}{%
\begin{tabular}{llrrrrrr}
\toprule
Dataset & Method & Nodes & Max depth & Mean depth & Depth std. & Branching & Calls/leaf \\
\midrule
LoCoMo  & \ours{} Batch    & 317 & 3.80 & 3.09 & 0.24 & 8.91  & 1.95 \\
        & \ours{} Cosine   & 352 & 4.00 & 3.46 & 0.60 & 5.10  & 0.96 \\
        & \ours{} Pointwise & 315 & 3.40 & 3.01 & 0.20 & 9.23  & 3.97 \\
        & \textsc{MemTree} & 474 & 5.90 & 3.82 & 0.92 & 2.45  & 3.67 \\
        & \textsc{RAPTOR}  & 290 & 2.20 & 2.13 & 0.08 & 34.37 & --   \\
\midrule
LongMemEval & \ours{} Batch    & 780  & 4.00 & 3.12 & 0.33 & 5.70  & 1.88 \\
            & \ours{} Cosine   & 833  & 4.00 & 3.36 & 0.48 & 4.38  & 0.93 \\
            & \ours{} Pointwise & 797  & 4.00 & 3.20 & 0.40 & 5.20  & 4.13 \\
            & \textsc{MemTree} & 1102 & 6.80 & 3.86 & 1.11 & 2.40  & 4.13 \\
            & \textsc{RAPTOR}  & 703  & 5.20 & 4.23 & 0.70 & 12.02 & --   \\
\midrule
RealMem & \ours{} Batch    & 947  & 4.00 & 3.46 & 0.49 & 3.81  & 1.94 \\
        & \ours{} Cosine   & 865  & 3.00 & 2.88 & 0.32 & 5.26  & 0.94 \\
        & \ours{} Pointwise & 835  & 3.00 & 2.99 & 0.08 & 6.55  & 3.93 \\
        & \textsc{MemTree} & 1190 & 6.90 & 3.65 & 1.20 & 2.42  & 4.09 \\
        & \textsc{RAPTOR}  & 734  & 4.40 & 3.66 & 0.65 & 21.54 & --   \\
\bottomrule
\end{tabular}%
}
\captionsetup{skip=10pt}
\caption{Cross-dataset tree-shape statistics, averaged over conversations in each dataset. LoCoMo has 10 conversations with 281 leaves on average, LongMemEval has 5 conversations with 643 leaves on average, and RealMem has 10 conversations with 698 leaves on average. Nodes denotes the per-tree node count. Max depth, mean depth, and depth std. summarize leaf-depth statistics. Branching denotes the mean branching factor of internal nodes. Calls/leaf denotes LLM construction calls per leaf. \textsc{RAPTOR} is omitted from Calls/leaf because its recursive clustering procedure lacks an online per-leaf analogue.}
\label{tab:tree-cross-dataset}
\end{table}

\Autoref{tab:tree-cross-dataset} shows that \ours{} produces
well-behaved trees in practice. Maximum leaf depth stays at $3$--$4$ on
conversations of $281$--$698$ utterances, far below the $O(T)$ worst-case
ceiling of the degenerate cascading regime
(\appref{sec:appendix-adversarial}), and depth standard deviations
are roughly half \textsc{MemTree}'s across most settings.

Branching and cost place \ours{} between the two baselines on both axes.
\textsc{RAPTOR} forms broad clustered levels (branching up to $34$) and
\textsc{MemTree} forms narrow semantic hierarchies ($\approx 2.4$);
\ours{} sits in between at $4$--$9$. Construction cost shows the same
pattern: cosine mode ($\approx 0.95$ calls/leaf) is cheaper than any
baseline, batch LLM ($\approx 1.9$) adds one batched frontier-selection
call per insertion, and sequential LLM ($\approx 4$) matches
\textsc{MemTree}'s budget while producing shallower, more uniform trees.


\subsection{LoCoMo \texttt{conv-47} Tree Visualization}
\label{sec:appendix-conv47}

\subsubsection{Tree Comparison}

\begin{figure}[t]
\centering
\includegraphics[width=\linewidth]{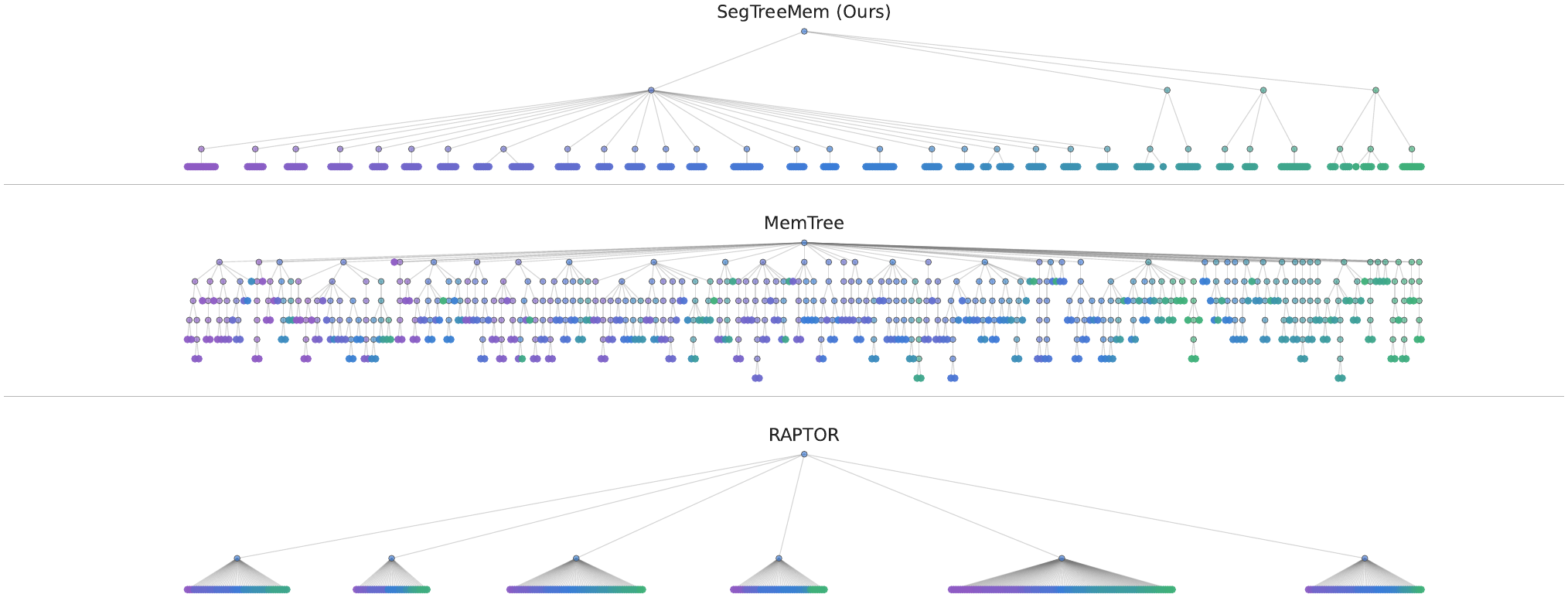}
\caption{Whole-tree visualizations of \ours{} (top), \textsc{MemTree}~\citep{memtree2024} (middle), and \textsc{RAPTOR}~\citep{raptor2024} (bottom) on LoCoMo \texttt{conv-47}, which spans $329$ utterances across $31$ dialogues. Each leaf (filled circle) is one utterance; internal nodes are drawn as outlined circles. Both leaves and internal nodes are colored by source dialogue index.}
\label{fig:appendix-trees-conv47}
\end{figure}

\texttt{conv-47} spans 31 sessions of conversation over approximately seven calendar months. The conversation is between two friends, James and John, and weaves together six recurring topics: gaming, programming, dogs, charity, travel, and romance. These topics are revisited across many distantly spaced sessions. The resulting tree structures for \ours{}, \textsc{MemTree}~\citep{memtree2024}, and \textsc{RAPTOR}~\citep{raptor2024} are shown in \Autoref{fig:appendix-trees-conv47}.

\newpage
\subsection{Adversarial Cases}
\label{sec:appendix-adversarial}

We analyze two adversarial input patterns for \ours{}'s online construction algorithm. 

\subsubsection{Maximally topic-switching input}
\label{sec:appendix-adversarial-comb}

\begin{figure}[h]
\centering
\includegraphics[width=0.55\linewidth]{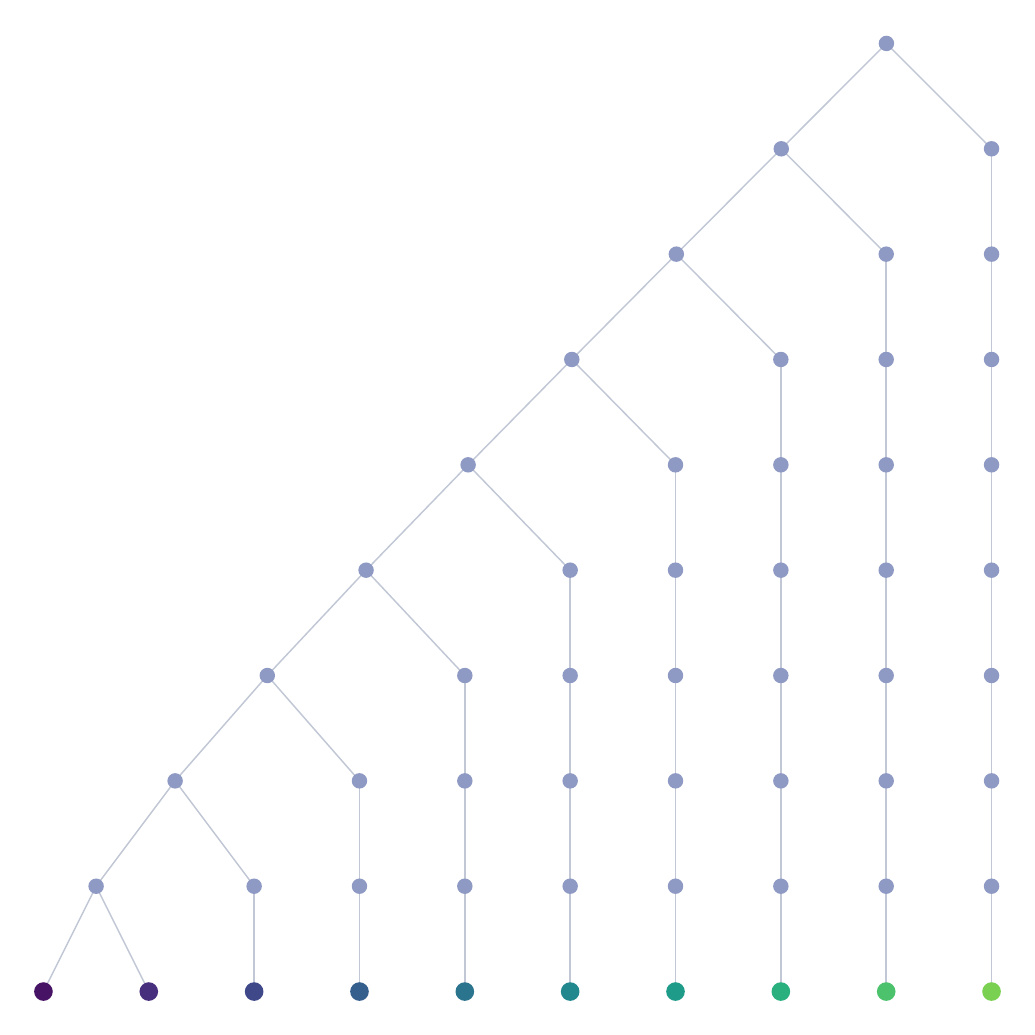}
\caption{Degenerate cascading segment tree produced by maximally topic-switching input. Each new utterance triggers a new root with a fresh right-side chain whose length grows with the height of the existing tree.}
\label{fig:appendix-adv-comb}
\end{figure}

The online insertion algorithm selects an attachment point from the
rightmost frontier, or creates a new root when no frontier node is
compatible with the incoming utterance. In the adversarial limit where
every new utterance is judged incompatible with every frontier node,
every insertion takes the new-root branch. After $T$ utterances, the
tree becomes a degenerate cascading segment tree
(\Autoref{fig:appendix-adv-comb}): each NONE step creates a new root
whose left subtree is the entire previous tree $T_t$ and whose right
subtree is a freshly created chain of $m-1$ internal nodes terminating
in the new leaf $z_{t+1}$. The resulting tree has height $O(T)$ and
$O(T^2)$ total nodes.

\paragraph{Consequences.}
This structure has two distinct costs. The first is storage: total node
count grows quadratically in $T$, and the $O(T^2)$ internal nodes each
require an annotation pass through $S$, making this regime strictly more
expensive than maintaining a flat list of utterances. The second is
representational. Each NONE step creates two kinds of internal nodes,
both representationally weak: the new root summarizes a long, incoherent
prefix ending at the latest utterance, while the $m-1$ chain nodes above
$z_{t+1}$ all carry the singleton interval $[t+1,t+1]$ and therefore
summarize a single utterance through redundant copies of one annotation.
Neither type captures focused conversational episodes. Structure-aware
propagation is correspondingly weakened: relevance signals between older
and newer leaves must pass through high-level ancestors that do not
correspond to meaningful temporal segments. Local leaf scoring still
retrieves individual utterances, but the tree structure itself
contributes little useful context.
\paragraph{When this arises.}
This case is a constructed worst case rather than a realistic operating
regime. It requires every incoming utterance to be judged
incompatible with every active temporal context on the frontier,
including the most recent leaf. Natural conversation does not satisfy
this condition: adjacent turns within a conversation almost always share some local discourse context. We do not observe degenerate cascading behavior on any
benchmark in this work; we describe it here to delimit the structural guarantees of the construction algorithm rather than to characterize typical inputs. The pattern would require either (i) an extremely  strict compatibility threshold or (ii) a synthetically constructed stream in which each utterance is drawn from a domain disjoint from every preceding utterance.

\subsubsection{Long-range topical recurrence}
\label{sec:appendix-adversarial-recurrent}

\begin{figure}[h]
\centering
\includegraphics[width=0.55\linewidth]{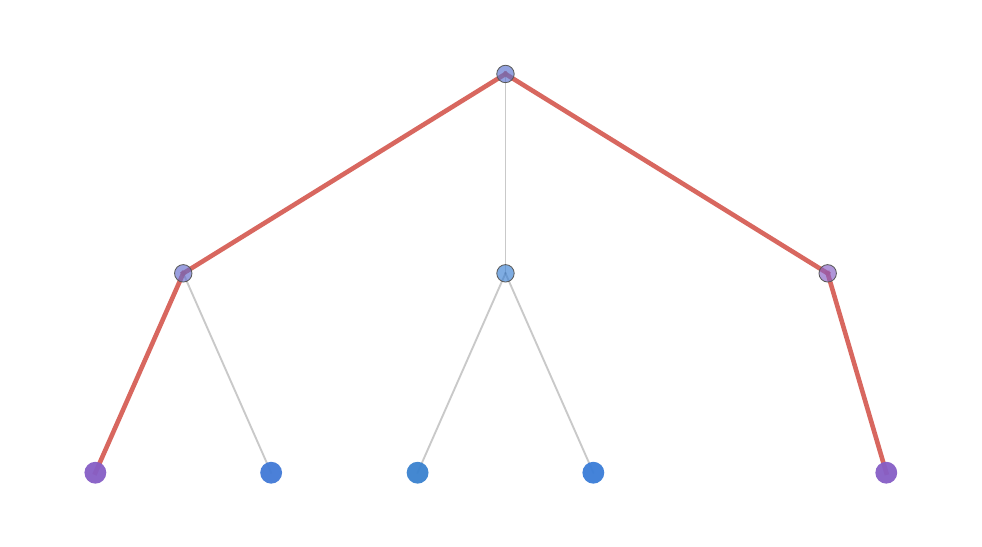}
\caption{Long-range topical recurrence. Two leaves on the same topic
(purple) are separated by an intervening span of utterances on other
topics (blue). Although the temporal segment-tree invariant is
preserved, the shortest tree path between the two topical mentions
passes through a high common ancestor. Finite-horizon propagation may
therefore fail to connect the two mentions.}
\label{fig:appendix-adv-recurrent}
\end{figure}

A second adversarial pattern occurs when the conversation repeatedly
returns to a previous topic after a long intervening span on other
topics. Because \ours{} inserts new utterances only through the rightmost
frontier, the later occurrence of the topic is placed according to its
current temporal context rather than being merged retroactively with the
earlier occurrence. As a result, two utterances about the same topic can
lie in disjoint subtrees whose lowest common ancestor covers a broad
intervening interval (\Autoref{fig:appendix-adv-recurrent}).

\paragraph{Consequences.}
The temporal segment-tree invariant still holds: every internal node
summarizes a contiguous span of the conversation. However, the recurrent
topic is split across multiple temporal regions, and no narrow internal
node summarizes all occurrences of that topic together. Since
structure-aware retrieval propagates scores for only a finite horizon
$H$, a relevance signal at one occurrence cannot necessarily reach the
other occurrence when their tree distance exceeds $H$. Increasing $H$
can reduce this issue, but also risks spreading relevance into unrelated
intervening content through high-level ancestors.

\paragraph{When this arises.}
Unlike maximally topic-switching input, this pattern is common in
long-horizon interaction. Users naturally return to ongoing projects,
people, preferences, locations, and periodic activities after many
unrelated utterances. The adversarial aspect is not the recurrence
itself, but the combination of recurrence with enough intervening
content that the related mentions become structurally distant in the
temporal tree.

\paragraph{Mitigation.}
The current retrieval framework partially mitigates this case through
local scoring and finite-horizon propagation. Collapsed retrieval is
robust when both recurrent mentions have annotations that independently
match the query, since it scores nodes without relying on tree distance.
Bottom-up propagation can also promote a relevant leaf to its local
temporal context, helping retrieve surrounding evidence from the same
episode. However, pure temporal propagation does not guarantee that a
single high-scoring occurrence will activate a distant recurrence of the
same topic. This is a genuine limitation of a purely temporal segment
tree, and suggests a natural extension: augment the temporal tree with
semantic cross-links between distant but related nodes, while preserving
the segment tree as the primary chronological scaffold.

\subsection{QA Case Analyses}
\label{sec:appendix-qa-cases}

We provide qualitative QA cases to illustrate how temporal tree
construction and score propagation affect downstream retrieval. We first
show two successful cases from the late-2022 portion of
\texttt{conv-47}, where \ours{} keeps the gold utterances and their
nearby context within the same local temporal segment. We then analyze a
representative failure of \ours{}, followed by two cases that illustrate
how score propagation changes the retrieved evidence.

\paragraph{Successful QA cases.}

The first two cases illustrate two ways that temporal segmentation helps
\ours{} retrieve the required evidence. In Q1, a narrow parent summary
preserves a local detail that is lost when gaming-related utterances are
aggregated more broadly. In Q2, a parent summary packages information
from adjacent sibling utterances, allowing retrieval to recover a short
answer even when the exact answer-bearing utterance is not retrieved
directly.

\begin{itemize}
\item \textbf{SM1, Local-detail preservation.}
When several adjacent utterances discuss the same event, \ours{} can
summarize them within a narrow temporal segment. This keeps specific
details attached to their local context rather than smoothing them into
a broad topic summary.

\item \textbf{SM2, Summary-mediated sibling evidence.}
When the answer is contained in a short neighboring utterance, the
parent summary can carry that detail into the retrieved evidence even if
retrieval lands on an adjacent utterance or parent node from the same
local temporal segment.
\end{itemize}

\paragraph{Q1 \textbar\ SM1: narrow temporal summaries preserve local details.}
\textbf{Question.} What did John suggest James practice before playing FIFA~23 together?\\
\textbf{Gold answer.} Control with a gamepad and timing.\\
\textbf{Relevant utterances} (5~Nov~2022).\quad
\textit{User 1:} Great idea! I hope it's easy to control.\quad
\textit{User 2:} Not at all, all you need is a gamepad and a
sense of timing.\quad
\textit{User 1:} Great! Well, I'll go train!

\compare{%
\begin{mem2cell}\small\textbf{\ours{}.} The depth-3 internal node
covering the 5~November dialogue encodes the answer in its summary:
``\dots John advised him to practice first \dots using a gamepad and
good timing.'' Because the FIFA-related utterances are grouped within a
narrow temporal segment, the gamepad/timing fact is preserved in the
summary seen by the retriever.
\end{mem2cell}%
}{%
\begin{mtcell}\small\textbf{\textsc{MemTree} and \textsc{RAPTOR}.}
\textsc{MemTree}'s retrieval is dominated by broad video-game clusters
that mix gaming utterances from different dates. These clusters recover
the topic of FIFA/gaming, but their summaries do not preserve the
specific gamepad/timing detail from 5~November. \textsc{RAPTOR}
retrieves nearby FIFA evidence, including the utterance where John says
James should ``practice a little first,'' but not the answer-bearing
detail about using a gamepad and having a sense of timing. Its generated
answer is therefore underspecified: James should ``practice a little
first.''
\end{mtcell}%
}

This case shows that temporal segmentation can preserve the granularity
needed for exact-answer questions. The baselines retrieve the correct
broad topic, but the retrieved evidence does not contain the specific
detail required by the question.

\begin{figure}[t]
\centering
\includegraphics[width=\linewidth]{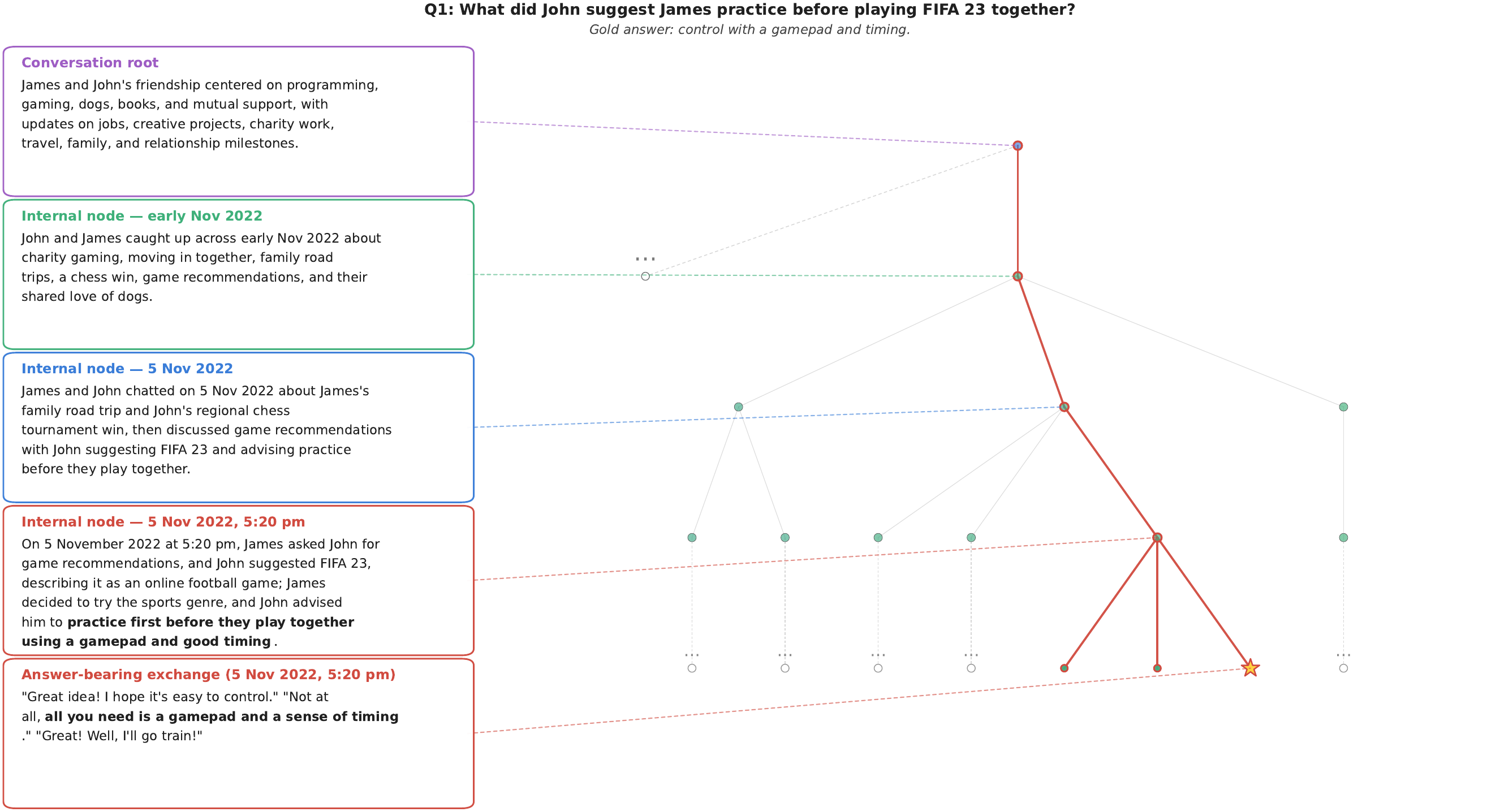}
\caption{Q1 ground-truth answer-bearing utterance and ancestor nodes in the LoCoMo \texttt{conv-47} \ours{} tree.}
\label{fig:appendix-q1-fifa}
\end{figure}

\paragraph{Q2 \textbar\ SM2: parent summaries recover adjacent sibling details.}
\textbf{Question.} What is the name of John's cousin's dog?\\
\textbf{Gold answer.} Luna.\\
\textbf{Relevant utterances} (7~Nov~2022).\quad
\textit{User 1:} This pup is so adorable! What's their name?\quad
\textit{User 2:} Their name is Luna.

\compare{%
\begin{mem2cell}\small\textbf{\ours{}.} The depth-3 internal node
covering the 7~November dialogue ends its summary with ``\dots
including John's cousin's dog Luna,'' so the parent summary already
contains the answer. The answer is therefore available through the
local parent node even when retrieval does not land directly on the
short answer-bearing utterance, ``Their name is Luna.''
\end{mem2cell}%
}{%
\begin{mtcell}\small\textbf{\textsc{MemTree} and \textsc{RAPTOR}.}
Both baselines retrieve nearby 7~November dog-related evidence in which
John shares a photo of his cousin's dog, but the retrieved evidence does
not name the dog. The exact name appears in an adjacent utterance, and
the retrieved evidence does not package that sibling fact with the
retrieved node. Thus, the baselines reach the right local neighborhood
but miss the short answer-bearing detail.
\end{mtcell}%
}

Together, Q1 and Q2 show two complementary benefits of temporal
segmentation. First, narrow temporal summaries can preserve local
details that may be smoothed away by broad semantic aggregation, as in
Q1. Second, parent summaries can package information from adjacent
sibling utterances, as in Q2. The latter case is especially important
for short answers such as names, where the answer-bearing utterance may
be semantically close to the retrieved evidence but absent from the
exact node text shown to the answer model.

\paragraph{Failure mode of \ours{}: missed details in internal node summaries.}

Although \ours{} improves retrieval by preserving chronological
structure, its summaries can still compress away exact local details.
This is most visible when the question requires an answer such as a date
or name. Score propagation can sometimes recover such details by moving
relevance from a relevant parent summary to its children, but it does
not guarantee that the answer-bearing utterance receives enough mass to
enter the evidence budget.

\paragraph{Q3: Internal summaries can omit exact local details.}
\textbf{Question.} When did John resume playing drums in his adulthood?\\
\textbf{Gold answer.} February 2022.

\compare{%
\begin{mem2cell}\small\textbf{\ours{}.} Retrieval reaches the correct
local temporal region, but the highest-ranked evidence from that region
is not the answer-bearing utterance. The retrieved parent summary
mentions that John resumed hobbies, but it does not preserve the
specific drum-playing fact needed to answer the question. Thus, the
temporal tree locates the right region of the conversation, while the
retrieved summary is not specific enough for literal date recall.
\end{mem2cell}%
}{%
\begin{mtcell}\small\textbf{\textsc{MemTree} and \textsc{RAPTOR}.}
\textsc{MemTree}'s leaf-level retrieval lands directly on the literal
drum mention. \textsc{RAPTOR} retrieves related drum evidence, including
a March utterance about John playing drums and a later utterance about
having played drums when he was younger, but it does not surface the
exact February resume-date evidence. This illustrates that literal
recall can depend sensitively on whether the answer-bearing utterance
itself enters the evidence budget.
\end{mtcell}%
}

This case illustrates a limitation of \ours{}: answer quality still
depends on whether the relevant detail is preserved in the retrieved
node annotations. This motivates either more detail-preserving
summarization prompts or a fallback to exact utterance-level evidence
when the query appears to require literal recall.

\paragraph{Score propagation.}

We next inspect two cases where score propagation changes the ranking or
composition of the retrieved evidence relative to collapsed retrieval.
These examples illustrate the intended behavior of propagation: moving
score toward nodes whose local temporal context also matches the query.

\paragraph{Q4: Propagation improves evidence selection.}
\textbf{Question.} What is John planning to do after receiving Samantha's phone number?\\
\textbf{Gold answer.} Call her.

Under collapsed retrieval, an unrelated programming/video-card utterance
is ranked slightly above a Samantha-related utterance. With
structure-aware propagation, the Samantha-related evidence is promoted
because its parent and ancestor nodes also describe the same local
Samantha-related temporal context. This small ranking change moves the
answer-bearing context above an unrelated near-tie.

\compare{%
\begin{mem2cell}\small\textbf{\ours{} with propagation.} Propagation
raises the score of the Samantha-related utterance by incorporating
support from its local parent context. The resulting evidence supports
the answer that John plans to call Samantha.
\end{mem2cell}%
}{%
\begin{mtcell}\small\textbf{Collapsed retrieval.} Without propagation,
the retrieval order leaves an unrelated programming/video-card utterance
above the Samantha-related evidence. The answer model therefore produces
a confused response instead of directly answering the intended event.
\end{mtcell}%
}

This case shows the intended use of propagation: the local relevance
score identifies candidate utterances, while the tree context helps
distinguish an utterance embedded in the relevant temporal segment from
a near-tie embedded in an unrelated segment. It also illustrates that
this is a propagation-specific effect rather than a general failure of
collapsed retrieval: \textsc{RAPTOR}'s collapsed retrieval surfaces the
literal Samantha phone-number evidence and answers this question
correctly.

\paragraph{Q5: Propagation can introduce distractors.}
\textbf{Question.} Which of James's family members have visited him in the last year?\\
\textbf{Gold answer.} Mother and sister.

This case shows the precision/recall tradeoff between propagation
policies. A broader top-down propagation policy admits a cousin-related
utterance because it shares family vocabulary with the query. However,
that extra evidence introduces a distractor, and the answer model
incorrectly adds cousins to the final answer. A tighter bottom-up policy
keeps retrieval concentrated around the directly relevant temporal
regions and answers with only mother and sister.

\compare{%
\begin{mem2cell}\small\textbf{Bottom-up propagation.} The bottom-up
policy keeps relevance concentrated around the utterances and parent
nodes that directly support the visit facts. The retrieved evidence
remains focused on the mother and sister mentions, producing the correct
answer.
\end{mem2cell}%
}{%
\begin{mtcell}\small\textbf{Top-down propagation.} The top-down policy
admits an incorrect cousin-related utterance. Although this evidence is
topically related to family, it is not part of the gold answer, and the
answer model over-includes cousins.
\end{mtcell}%
}

Together, Q4 and Q5 show that propagation is useful when structural
context agrees with the query, but can hurt when nearby or descendant
nodes contain semantically related distractors. This supports the main
experimental finding that propagation should be used with limited
horizons and moderate strength: it is a retrieval-time correction that
can improve evidence selection, not a substitute for local node
relevance.
\newpage
\section{Additional Results}

\subsection{Qwen Results}
\label{sec:appendix-qwen-main-results}

\paragraph{Qwen backbone results.}
\Autoref{tab:qwen-main-results} reports the same main comparison when
LLM-dependent memory construction and answer generation use
\texttt{qwen3.5-flash}. \ours{} remains the strongest method under this
backbone on all three benchmarks. The no-propagation variant already improves
over the strongest external baseline in primary score on LoCoMo,
LongMemEval-MAB, and RealMem, showing that the temporally ordered tree structure
itself contributes beyond the answer model. The Best variant further improves
over no propagation by \(2.4\) points on LoCoMo, \(5.1\) points on
LongMemEval-MAB, and \(4.2\) points on RealMem, indicating that
structure-aware propagation continues to provide useful context expansion under
a different LLM backbone.

\paragraph{Effect of propagation under Qwen.}
The gains from Best over no propagation are not uniform across metrics. On
LoCoMo and RealMem, the Best variant improves both the primary judge score and
token-level F1, suggesting that propagation helps retrieve evidence that is
both more judge-sufficient and more lexically aligned with the reference
answers. On LongMemEval-MAB, the primary accuracy gain is large, while F1
changes only modestly. This pattern is consistent with LongMemEval-MAB's
answer format: many questions are judged by whether the answer state is
correct, and the token-level overlap metric is less sensitive to whether the
retrieved evidence supports the correct update or temporal relation. Thus, the
Qwen results support the same conclusion as the main results: propagation is
most useful as a mechanism for selecting the appropriate temporal context, not
merely as a way to increase lexical overlap.

\definecolor{addlightpurple}{RGB}{244,238,255}
\definecolor{addheaderpurple}{RGB}{226,214,245}
\definecolor{addcigray}{RGB}{95,95,95}
\definecolor{addgroupgray}{RGB}{80,80,80}

\newcommand{\addscoreci}[2]{#1 {\scriptsize\textcolor{addcigray}{[#2]}}}
\newcommand{\addbestscoreci}[2]{\textbf{#1} {\scriptsize\textcolor{addcigray}{[#2]}}}
\newcommand{\addbest}[1]{\textbf{#1}}
\newcommand{\addmethodgroupmain}[1]{%
  \multicolumn{8}{@{}l}{\textit{\textcolor{addgroupgray}{#1}}}%
}
\newcommand{\addmethodgrouplocomo}[1]{%
  \multicolumn{13}{@{}l}{\textit{\textcolor{addgroupgray}{#1}}}%
}
\newcommand{\addmethodgrouplme}[1]{%
  \multicolumn{15}{@{}l}{\textit{\textcolor{addgroupgray}{#1}}}%
}

\subsubsection{Propagation Sweep under Qwen}
\label{sec:appendix-qwen-propagation-sweep}

\begin{figure*}[t]
    \centering
    \includegraphics[width=\textwidth]{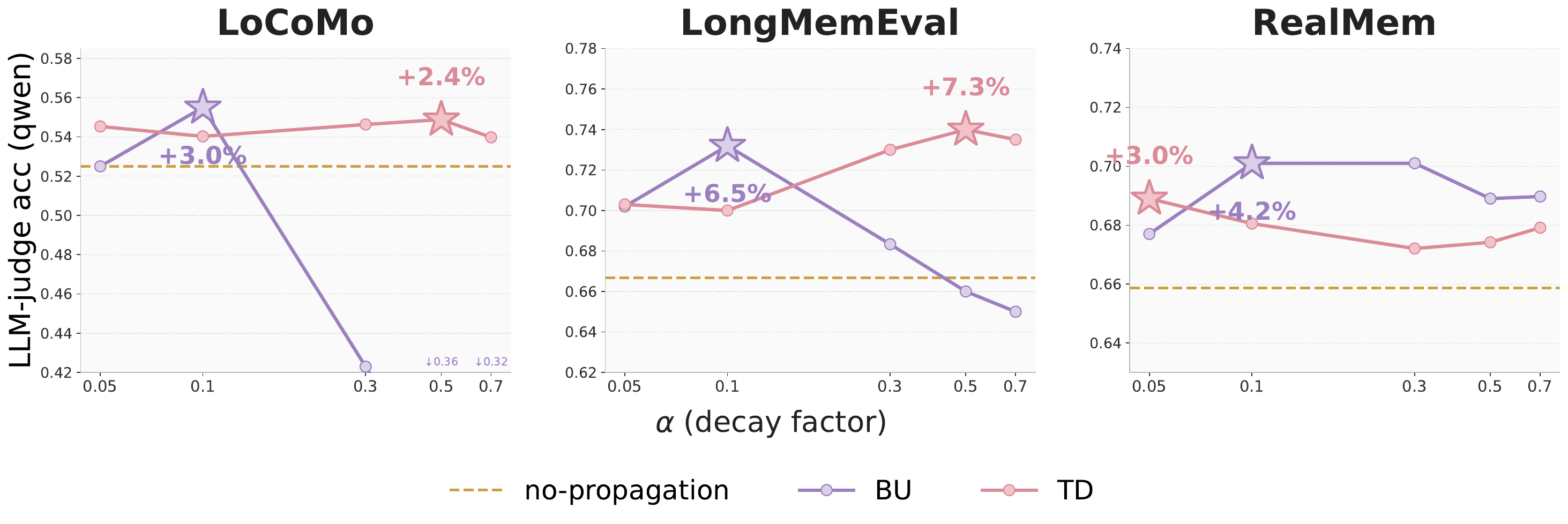}
    \caption{Effect of propagation direction and decay factor under the
    \texttt{qwen3.5-flash} backbone. The dashed horizontal line denotes
    no-propagation retrieval. BU denotes bottom-up propagation and TD denotes
    top-down propagation. Stars mark the best setting within each propagation
    direction, and percentage annotations report the improvement over the
    no-propagation setting. The plotted score is LLM-judge accuracy for LoCoMo
    and LongMemEval-MAB, and perfect-rate for RealMem.}
    \label{fig:qwen-propagation-sweep}
\end{figure*}

\paragraph{Propagation remains beneficial under Qwen.}
\Autoref{fig:qwen-propagation-sweep} shows that structure-aware retrieval also
improves \ours{} when LLM-dependent memory construction and answer generation
use \texttt{qwen3.5-flash}. The no-propagation baseline is already strong, but
both propagation directions produce additional gains on at least one setting for
each benchmark. This supports the same conclusion as the main GPT-backed
results: the tree structure is useful by itself, and propagation further improves
retrieval by changing how relevance is allocated across the temporal hierarchy.

\paragraph{The best propagation direction is benchmark-dependent.}
The Qwen sweep again shows that there is no universally optimal propagation
policy. On LoCoMo, top-down propagation gives a stable improvement at moderate
decay, while bottom-up propagation is useful only at small decay and becomes
unstable when too much mass is pushed upward. On LongMemEval-MAB, both
directions improve over no propagation, with top-down propagation giving the
largest gain at a moderate decay factor. On RealMem, bottom-up propagation gives
the strongest result, consistent with the intuition that RealMem often requires
recovering broader project- or task-level state rather than only a single
answer-bearing exchange.

\paragraph{Decay controls the granularity--drift tradeoff.}
The shape of the curves illustrates the role of \(\alpha\). Small and moderate
values of \(\alpha\) allow retrieval to incorporate structurally related context
while keeping the initial semantic match anchored. Large values can move too
much probability mass away from the answer-bearing node. This effect is most
visible for bottom-up propagation on LoCoMo, where aggressive upward expansion
substantially degrades accuracy. The same pattern appears more mildly on
LongMemEval-MAB, where bottom-up propagation improves at small decay but falls
below the best top-down setting as decay increases.

\paragraph{Backbone robustness.}
Together with \Autoref{tab:qwen-main-results}, this sweep shows that the
propagation behavior is not an artifact of the GPT-family backbone. The absolute
scores differ across answer models, and Qwen has a distinct category-level
failure profile, especially on LoCoMo adversarial questions. Nevertheless, the
same high-level retrieval pattern persists: useful propagation is task- and
granularity-dependent. Detail-oriented settings favor top-down or weak
propagation, whereas state-oriented settings such as RealMem benefit more from
bottom-up context expansion.

\begin{table*}[t!]
\centering
\small
\setlength{\tabcolsep}{2.8pt}
\renewcommand{\arraystretch}{1.10}
\resizebox{\textwidth}{!}{%
\begin{tabular}{@{}llcccccc@{}}
\toprule
\rowcolor{addheaderpurple}
\textbf{Method} &
\textbf{Variant} &
\multicolumn{2}{c}{\textbf{LoCoMo}} &
\multicolumn{2}{c}{\textbf{LongMemEval-MAB}} &
\multicolumn{2}{c}{\textbf{RealMem}} \\
\rowcolor{addheaderpurple}
&
&
\textbf{LLM} & \textbf{F1} &
\textbf{LLM} & \textbf{F1} &
\textbf{LLM} & \textbf{F1} \\
\midrule

\addmethodgroupmain{Retrieval-only memory baselines} \\

BM25 \citep{bm251994}
& Default
& \addscoreci{0.469}{0.447, 0.491}
& \addscoreci{0.157}{0.150, 0.164}
& \addscoreci{0.542}{0.485, 0.598}
& \addscoreci{0.137}{0.121, 0.154}
& \addscoreci{0.611}{0.585, 0.636}
& \addscoreci{0.359}{0.353, 0.364} \\

Dense \citep{dense2020}
& Default
& \addscoreci{0.477}{0.455, 0.499}
& \addscoreci{0.153}{0.147, 0.160}
& \addscoreci{0.590}{0.534, 0.646}
& \addscoreci{0.140}{0.125, 0.156}
& \addscoreci{0.631}{0.606, 0.656}
& \addscoreci{0.363}{0.357, 0.368} \\

\midrule

\addmethodgroupmain{Graph-structured memory baselines} \\

A-MEM \citep{amem2025}
& Default
& \addscoreci{0.496}{0.474, 0.518}
& \addscoreci{0.155}{0.149, 0.162}
& \addscoreci{0.599}{0.543, 0.654}
& \addscoreci{0.139}{0.123, 0.154}
& \addscoreci{0.635}{0.610, 0.660}
& \addscoreci{0.361}{0.355, 0.366} \\

Mem0 \citep{mem02025}
& Default
& \addscoreci{0.354}{0.332, 0.375}
& \addscoreci{0.128}{0.122, 0.133}
& \addscoreci{0.517}{0.460, 0.573}
& \addbestscoreci{0.147}{0.132, 0.162}
& \addscoreci{0.486}{0.460, 0.512}
& \addscoreci{0.351}{0.346, 0.356} \\

HippoRAG \citep{hipporag2024}
& Default
& \addscoreci{0.470}{0.448, 0.492}
& \addscoreci{0.151}{0.145, 0.158}
& \addscoreci{0.612}{0.557, 0.667}
& \addscoreci{0.142}{0.126, 0.159}
& \addscoreci{0.629}{0.603, 0.654}
& \addscoreci{0.359}{0.353, 0.364} \\

\midrule

\addmethodgroupmain{Tree-structured memory baselines} \\

RAPTOR \citep{raptor2024}
& Default
& \addscoreci{0.518}{0.496, 0.540}
& \addscoreci{0.146}{0.140, 0.152}
& \addscoreci{0.593}{0.538, 0.649}
& \addscoreci{0.140}{0.124, 0.156}
& \addscoreci{0.618}{0.592, 0.643}
& \addscoreci{0.362}{0.356, 0.367} \\

MemTree \citep{memtree2024}
& Default
& \addscoreci{0.369}{0.347, 0.390}
& \addscoreci{0.118}{0.112, 0.123}
& \addscoreci{0.543}{0.487, 0.600}
& \addscoreci{0.137}{0.121, 0.152}
& \addscoreci{0.597}{0.571, 0.622}
& \addscoreci{0.360}{0.355, 0.365} \\

\midrule

\addmethodgroupmain{Proposed tree-structured memory method} \\

\rowcolor{addlightpurple}
\textbf{\ours}
& No-Propagation
& \addscoreci{0.525}{0.503, 0.547}
& \addscoreci{0.161}{0.154, 0.167}
& \addscoreci{0.669}{0.618, 0.720}
& \addscoreci{0.128}{0.112, 0.143}
& \addscoreci{0.659}{0.634, 0.683}
& \addscoreci{0.359}{0.354, 0.365} \\

\rowcolor{addlightpurple}
\textbf{\ours}
& Best
& \addbestscoreci{0.549}{0.527, 0.571}
& \addbestscoreci{0.168}{0.162, 0.175}
& \addbestscoreci{0.720}{0.669, 0.771}
& \addscoreci{0.144}{0.129, 0.159}
& \addbestscoreci{0.701}{0.677, 0.724}
& \addbestscoreci{0.364}{0.358, 0.369} \\

\bottomrule
\end{tabular}%
}
\caption{Qwen answer-quality results across long-horizon memory benchmarks.
LLM denotes LLM-judge accuracy and F1 denotes token-level F1. Each reported score includes its
95\% confidence interval in brackets. For \ours{}, Best denotes the best propagation
setting selected by LLM-judge accuracy from the retrieval sweep.
All LLM-dependent memory construction and answer generation use
\texttt{qwen3.5-flash}. Bold values mark the best score in each metric column.}
\label{tab:qwen-main-results}
\end{table*}

\newpage

\subsection{Question-Type Breakdown}
\label{sec:appendix-question-type-breakdown}

\subsubsection{LoCoMo Breakdown}
\Autoref{tab:locomo-breakdown-gpt-qwen} shows that \ours{} obtains the best
aggregate LoCoMo accuracy under both \texttt{gpt-5.4-mini} and
\texttt{qwen3.5-flash}, but the category-level behavior differs substantially
between the two backbones. Under \texttt{gpt-5.4-mini}, \ours{} Best leads on
single-hop, multi-hop, temporal, open-domain, and overall accuracy. The only
category where it does not lead is adversarial QA, where A-MEM, RAPTOR, and
HippoRAG remain stronger. This indicates that temporal hierarchy and
propagation are especially helpful for locating answer-bearing evidence across
ordinary single-hop, multi-hop, and temporal questions, while adversarial
questions additionally require robust rejection of unsupported premises.

\paragraph{Backbone sensitivity on LoCoMo.}
The \texttt{qwen3.5-flash} LoCoMo results show a different error profile. The
largest degradation is concentrated in adversarial questions: all structured
methods drop sharply on this category, and the strongest adversarial score under
Qwen comes from BM25 rather than from a structured memory method. Despite this
adversarial weakness, \ours{} Best remains the best aggregate method because it
retains strong performance on multi-hop, temporal, and open-domain questions.
This suggests that the retrieval structure continues to help recover relevant
evidence, but the answer backbone affects whether the model correctly refuses
or qualifies answers when the question contains a misleading premise.


\begin{table*}[t]
\centering
\scriptsize
\setlength{\tabcolsep}{1.8pt}
\renewcommand{\arraystretch}{1.08}
\resizebox{\textwidth}{!}{%
\begin{tabular}{@{}lrrrrrr!{\hspace{0.7em}\vrule\hspace{0.7em}}rrrrrr@{}}
\toprule
\rowcolor{addheaderpurple}
\textbf{Method} &
\multicolumn{6}{c}{\textbf{\texttt{gpt-5.4-mini}}} &
\multicolumn{6}{c}{\textbf{\texttt{qwen3.5-flash}}} \\
\cmidrule(lr){2-7}\cmidrule(lr){8-13}
\rowcolor{addheaderpurple}
&
\textbf{Single-hop} &
\textbf{Multi-hop} &
\textbf{Temporal} &
\textbf{Open-domain} &
\textbf{Adversarial} &
\textbf{All} &
\textbf{Single-hop} &
\textbf{Multi-hop} &
\textbf{Temporal} &
\textbf{Open-domain} &
\textbf{Adversarial} &
\textbf{All} \\
\midrule

\addmethodgrouplocomo{Retrieval-only memory baselines} \\

BM25 \citep{bm251994}
& 0.252 & 0.389 & 0.354 & 0.734 & 0.191 & 0.469
& 0.291 & 0.442 & 0.302 & 0.707 & \addbest{0.186} & 0.469 \\

Dense \citep{dense2020}
& 0.340 & 0.421 & 0.385 & 0.750 & 0.168 & 0.490
& 0.383 & 0.433 & 0.354 & 0.717 & 0.143 & 0.477 \\

\midrule

\addmethodgrouplocomo{Graph-structured memory baselines} \\

A-MEM \citep{amem2025}
& 0.422 & 0.424 & 0.417 & 0.779 & \addbest{0.576} & 0.608
& 0.429 & 0.470 & 0.385 & 0.737 & 0.126 & 0.496 \\

Mem0 \citep{mem02025}
& 0.191 & 0.324 & 0.323 & 0.597 & 0.471 & 0.454
& 0.206 & 0.405 & 0.260 & 0.507 & 0.141 & 0.353 \\

HippoRAG \citep{hipporag2024}
& 0.372 & 0.402 & 0.375 & 0.725 & 0.565 & 0.570
& 0.394 & 0.417 & 0.375 & 0.711 & 0.121 & 0.470 \\

\midrule

\addmethodgrouplocomo{Tree-structured memory baselines} \\

RAPTOR \citep{raptor2024}
& 0.482 & 0.442 & 0.396 & 0.767 & 0.567 & 0.593
& \addbest{0.567} & 0.421 & 0.375 & 0.769 & 0.114 & 0.518 \\

MemTree \citep{memtree2024}
& 0.291 & 0.327 & 0.406 & 0.630 & 0.455 & 0.483
& 0.294 & 0.283 & 0.312 & 0.548 & 0.150 & 0.369 \\

\midrule

\addmethodgrouplocomo{Proposed tree-structured memory method} \\

\rowcolor{addlightpurple}
\textbf{\ours} No-Propagation
& 0.470 & 0.554 & 0.385 & 0.879 & 0.410 & 0.639
& 0.445 & 0.519 & 0.288 & 0.783 & 0.067 & 0.525 \\

\rowcolor{addlightpurple}
\textbf{\ours} Best
& \addbest{0.526} & \addbest{0.580} & \addbest{0.542} & \addbest{0.894} & 0.420 & \addbest{0.668}
& 0.514 & \addbest{0.558} & \addbest{0.458} & \addbest{0.811} & 0.090 & \addbest{0.549} \\

\bottomrule
\end{tabular}%
}
\caption{LoCoMo question-type LLM-judge accuracy under two answer and
LLM-dependent memory-construction backbones. The five LoCoMo categories are
single-hop, multi-hop, temporal, open-domain, and adversarial. For \ours{},
Identity denotes no propagation, and Best denotes the best propagation setting
under the corresponding backbone. Bold values mark the best score in each
category within each backbone.}
\label{tab:locomo-breakdown-gpt-qwen}
\end{table*}

\subsubsection{LongMemEval-MAB Breakdown}
\Autoref{tab:lme-breakdown-gpt-qwen} shows a cleaner pattern than LoCoMo.
Under both \texttt{gpt-5.4-mini} and \texttt{qwen3.5-flash}, \ours{} Best
achieves the highest aggregate accuracy. The strongest gains appear in
multi-session, single-session preference, and temporal-reasoning questions,
which are precisely the categories that require maintaining state across turns
or resolving information through temporal context. By contrast, the
single-session assistant category is nearly saturated for many methods, and
the single-session user and knowledge-update categories are sometimes led by
external baselines. Therefore, the aggregate improvement is not driven by easy
single-session cases; it comes primarily from categories where temporal memory
organization is expected to matter.

\paragraph{Propagation improves state-tracking categories.}
Comparing identity/no-propagation with Best shows that propagation is most
useful for categories that require moving between a local evidence match and a
broader memory state. On LongMemEval-MAB, Best improves over no propagation on
multi-session and temporal-reasoning questions under both backbones. The same
trend appears for preference questions, where the answer often depends on
recovering an accumulated or updated user state rather than a single isolated
utterance. These category-level gains explain why the aggregate Best row
improves over the identity row even when some single-session categories are
already near ceiling.

\paragraph{Backbone effects on LongMemEval-MAB.}
The \texttt{qwen3.5-flash} backbone raises LongMemEval-MAB accuracy for most
methods relative to the GPT-family runs, but the relative conclusion remains
unchanged: \ours{} Best is the strongest aggregate method, and its advantage is
largest on the temporally demanding categories. This is important because the
Qwen setting changes both LLM-dependent construction and answer generation.
The fact that the same category-level pattern persists suggests that the gains
are not an artifact of a particular answer model, but reflect the interaction
between temporally ordered memory construction and structure-aware retrieval.

\begin{table*}[t]
\centering
\scriptsize
\setlength{\tabcolsep}{1.4pt}
\renewcommand{\arraystretch}{1.08}
\resizebox{\textwidth}{!}{%
\begin{tabular}{@{}lrrrrrrr!{\hspace{0.7em}\vrule\hspace{0.7em}}rrrrrrr@{}}
\toprule
\rowcolor{addheaderpurple}
\textbf{Method} &
\multicolumn{7}{c}{\textbf{\texttt{gpt-5.4-mini}}} &
\multicolumn{7}{c}{\textbf{\texttt{qwen3.5-flash}}} \\
\cmidrule(lr){2-8}\cmidrule(lr){9-15}
\rowcolor{addheaderpurple}
&
\textbf{Know.} &
\textbf{Multi} &
\textbf{Asst.} &
\textbf{Pref.} &
\textbf{User} &
\textbf{Temp.} &
\textbf{All} &
\textbf{Know.} &
\textbf{Multi} &
\textbf{Asst.} &
\textbf{Pref.} &
\textbf{User} &
\textbf{Temp.} &
\textbf{All} \\
\midrule

\addmethodgrouplme{Retrieval-only memory baselines} \\

BM25 \citep{bm251994}
& 0.578 & 0.253 & 0.967 & 0.367 & \addbest{0.889} & 0.320 & 0.497
& 0.644 & 0.307 & \addbest{1.000} & 0.400 & \addbest{0.933} & 0.360 & 0.542 \\

Dense \citep{dense2020}
& 0.689 & 0.307 & 0.967 & 0.467 & \addbest{0.889} & 0.253 & 0.520
& 0.756 & 0.333 & \addbest{1.000} & 0.633 & 0.889 & 0.387 & 0.590 \\

\midrule

\addmethodgrouplme{Graph-structured memory baselines} \\

A-MEM \citep{amem2025}
& 0.711 & 0.387 & 0.967 & 0.600 & 0.822 & 0.280 & 0.553
& 0.733 & 0.427 & \addbest{1.000} & 0.633 & 0.911 & 0.333 & 0.599 \\

Mem0 \citep{mem02025}
& 0.756 & 0.400 & 0.767 & 0.433 & \addbest{0.889} & 0.280 & 0.537
& 0.711 & 0.373 & 0.800 & 0.333 & 0.889 & 0.280 & 0.517 \\

HippoRAG \citep{hipporag2024}
& \addbest{0.778} & 0.373 & 0.967 & 0.533 & 0.867 & 0.280 & 0.560
& \addbest{0.822} & 0.373 & \addbest{1.000} & 0.700 & 0.889 & 0.373 & 0.612 \\

\midrule

\addmethodgrouplme{Tree-structured memory baselines} \\

RAPTOR \citep{raptor2024}
& 0.644 & 0.333 & 0.967 & 0.533 & 0.867 & 0.253 & 0.523
& 0.800 & 0.347 & \addbest{1.000} & 0.667 & 0.889 & 0.347 & 0.593 \\

MemTree \citep{memtree2024}
& 0.556 & 0.227 & \addbest{1.000} & 0.633 & 0.822 & 0.227 & 0.483
& 0.644 & 0.320 & 0.967 & 0.633 & 0.889 & 0.293 & 0.543 \\

\midrule

\addmethodgrouplme{Proposed tree-structured memory method} \\

\rowcolor{addlightpurple}
\textbf{\ours} No-Propagation
& 0.632 & 0.437 & \addbest{1.000} & 0.652 & 0.821 & 0.536 & 0.630
& 0.629 & 0.406 & \addbest{1.000} & 0.654 & 0.796 & 0.545 & 0.669 \\

\rowcolor{addlightpurple}
\textbf{\ours} Best
& 0.718 & \addbest{0.488} & 0.973 & \addbest{0.702} & 0.829 & \addbest{0.595} & \addbest{0.670}
& 0.778 & \addbest{0.520} & \addbest{1.000} & \addbest{0.767} & 0.867 & \addbest{0.667} & \addbest{0.720} \\

\bottomrule
\end{tabular}%
}
\caption{LongMemEval-MAB question-type LLM-judge accuracy under two answer and
LLM-dependent memory-construction backbones. Columns are knowledge update
(Know.), multi-session (Multi), single-session assistant (Asst.),
single-session preference (Pref.), single-session user (User), temporal
reasoning (Temp.), and aggregate accuracy (All). For \ours{}, Identity denotes
no propagation, and Best denotes the best propagation setting under the
corresponding backbone. Bold values mark the best score in each category within
each backbone.}
\label{tab:lme-breakdown-gpt-qwen}
\end{table*}
\newpage
\subsection{Additional Propagation Analysis}
\label{sec:appendix-additional-propagation}

To complement the propagation sweep in \Autoref{fig:propagation_sweep}, we
provide two additional diagnostics for structure-aware retrieval. First, we
analyze how accuracy changes jointly with the propagation horizon \(H\) and
decay factor \(\alpha\). Second, we inspect how propagation changes the
temporal level of the retrieved evidence. Together, these analyses clarify why
propagation helps in some settings but hurts in others: the propagation policy
does not simply add more context, but changes the granularity at which evidence
is retrieved from the temporal hierarchy.

\begin{figure*}[t]
    \centering
    \includegraphics[width=\textwidth]{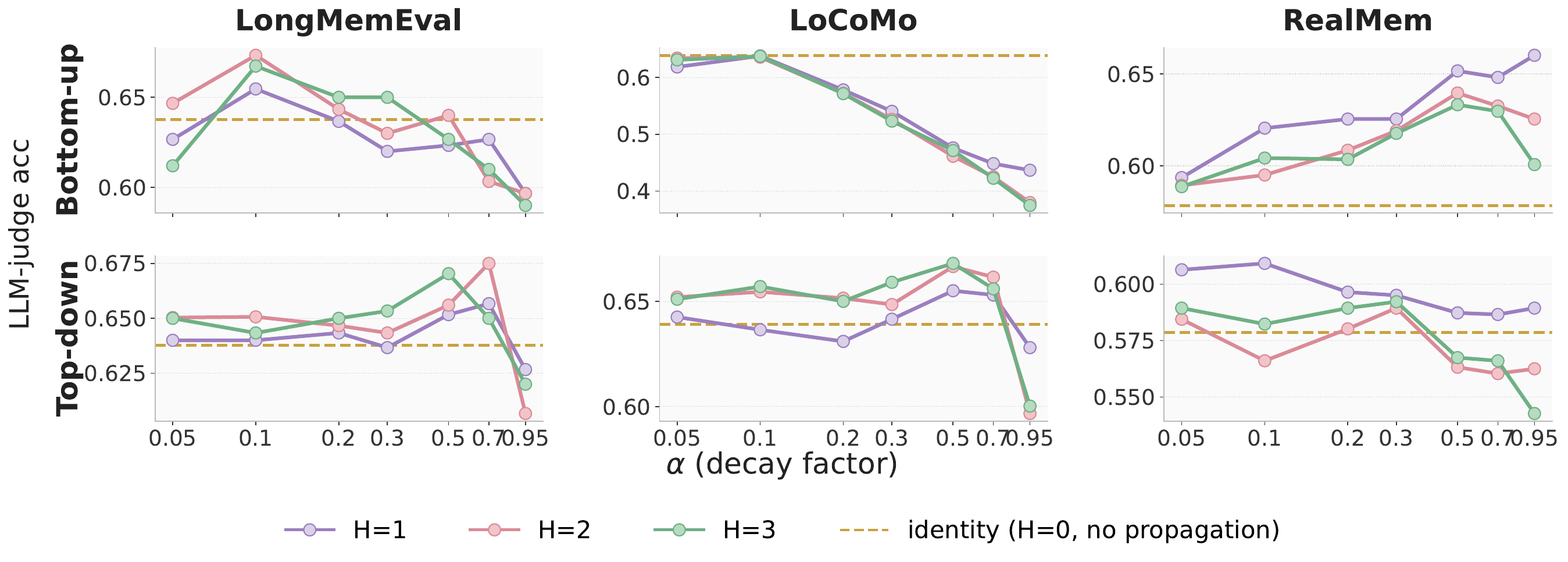}
    \caption{Accuracy as a function of the decay factor \(\alpha\) and
    propagation horizon \(H\), using batch-LLM \ours{} trees. The dashed
    horizontal line denotes the no-propagation baseline \(H=0\). Solid curves
    report finite-horizon bottom-up and top-down propagation with
    \(H\in\{1,2,3\}\) over the grid
    \(\alpha\in\{0.05,0.10,0.20,0.30,0.50,0.70,0.95\}\). The plotted score is
    LLM-judge accuracy for LongMemEval-MAB and LoCoMo, and perfect-rate for
    RealMem.}
    \label{fig:appendix-alpha-h-curves}
\end{figure*}

\paragraph{Propagation is most useful at short horizons.}
\Autoref{fig:appendix-alpha-h-curves} shows that one or two propagation hops are
usually sufficient to capture the useful structural context. Increasing the
horizon beyond this range rarely gives a consistent gain and can make retrieval
less stable, especially when paired with a large decay factor. This supports the
finite-horizon design used in the main experiments: relevant context is often
near the initial semantic match in the tree, either as a parent summary, a child
span, or a nearby hierarchical neighbor. Long-range propagation can move too
much probability mass away from the answer-bearing region.

\paragraph{Decay controls the amount of structural drift.}
The curves also show that \(\alpha\) is the main knob controlling the tradeoff
between semantic anchoring and structural expansion. Small and moderate
\(\alpha\) values allow the retriever to incorporate nearby hierarchical context
while preserving the original query match. Very large \(\alpha\) values can
overweight propagated mass and cause retrieval drift. This is most visible for
bottom-up propagation on LoCoMo, where increasing \(\alpha\) steadily lowers
accuracy across horizons. In contrast, LongMemEval-MAB and RealMem tolerate
moderate upward expansion better, because many of their questions benefit from
recovering broader state or temporally aggregated context.

\paragraph{The best direction depends on the benchmark.}
The horizon sweep confirms that there is no universally best propagation
direction. LoCoMo benefits more from top-down propagation at moderate decay,
which is consistent with its many detail-oriented questions: broad summaries can
identify the relevant episode, but the answer often requires retrieving a more
specific descendant. RealMem shows the opposite tendency. Its strongest gains
come from bottom-up propagation, suggesting that local matches are often useful
entry points but the final answer depends on broader project- or task-level
state. LongMemEval-MAB lies between these regimes: both directions can improve
over identity retrieval, but performance is sensitive to the chosen decay and
horizon.

\begin{figure*}[t]
    \centering
    \includegraphics[width=\textwidth]{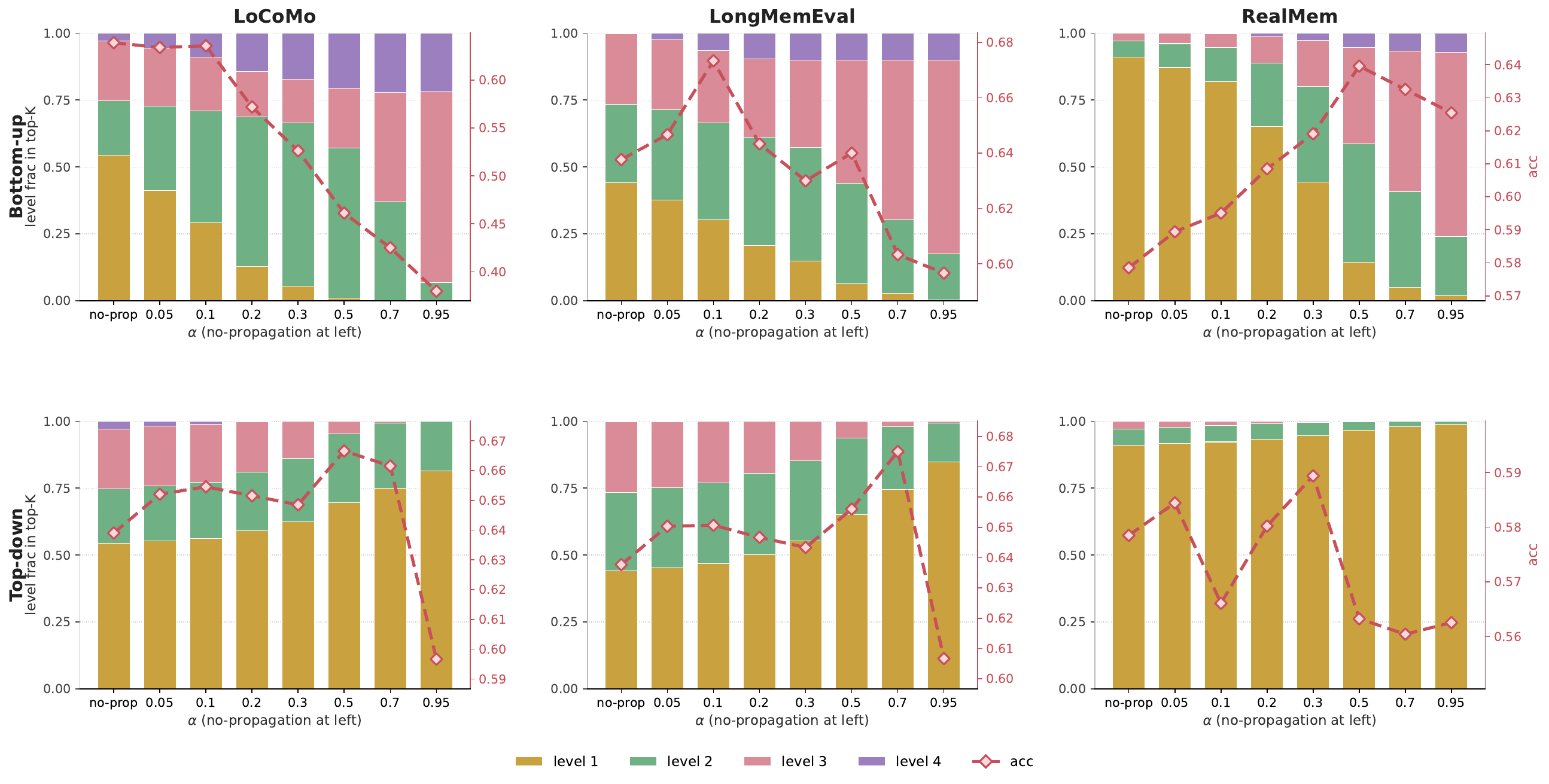}
    \caption{Retrieved-node level distribution as a function of the propagation
    setting, using batch-LLM \ours{} trees and a fixed propagation horizon
    \(H=2\). The leftmost bar in each panel is the identity baseline. Stacked
    bars show the fraction of retrieved nodes at each dataset-specific temporal
    level, and the black line shows the corresponding accuracy or perfect-rate
    score. Bottom-up propagation increasingly shifts retrieval toward broader
    internal nodes as \(\alpha\) grows, whereas top-down propagation remains
    more concentrated on exchange-level evidence.}
    \label{fig:appendix-retrieval-levels}
\end{figure*}

\paragraph{Top-down propagation remains exchange-oriented.}
\Autoref{fig:appendix-retrieval-levels} shows that top-down propagation keeps
most retrieved evidence close to the leaves. Even as \(\alpha\) increases, the
retrieved set remains dominated by exchange-level nodes, with only a moderate
increase in broader temporal spans at intermediate settings. This matches the
semantics of top-down propagation: relevance that initially matches a
higher-level summary is pushed downward toward finer-grained descendants. The
resulting retrieval behavior is well suited to questions that require localized
details, which helps explain why top-down propagation is competitive on LoCoMo
and LongMemEval-MAB.

\paragraph{Bottom-up propagation expands toward broader temporal context.}
Bottom-up propagation exhibits the complementary pattern. As \(\alpha\)
increases, retrieved evidence shifts from exchange-level nodes toward
higher-level temporal units, such as subsessions and sessions in
LongMemEval-MAB and LoCoMo, or broader episode-like segments in RealMem. This
behavior reflects the intended role of bottom-up propagation: a highly relevant
local match can promote its enclosing temporal context, allowing retrieval to
recover the broader state surrounding the matched exchange. This broader context
is helpful on RealMem, but it can hurt on LoCoMo when the question requires a
narrow answer-bearing exchange.

\paragraph{Granularity explains the dataset-dependent sweep.}
The two diagnostics together explain the propagation behavior in
\Autoref{fig:propagation_sweep}. On LoCoMo, stronger bottom-up propagation moves
retrieval away from exchange-level evidence and is accompanied by a clear drop
in accuracy, indicating that many questions require narrow, detail-preserving
evidence. On RealMem, moderate bottom-up propagation improves performance while
also increasing the share of broader internal nodes, suggesting that project- or
task-level context is often useful. LongMemEval-MAB lies between these extremes:
a small amount of structural expansion can help, but aggressive propagation can
dilute answer-bearing local evidence.

Overall, these results support the interpretation in
\Autoref{sec:results}: tree-aware retrieval is best understood as a
\emph{granularity-control mechanism}. The propagation direction determines
whether retrieval expands toward broader context or finer detail, while the
decay factor \(\alpha\) and horizon \(H\) control the strength and distance of
that expansion. The best setting therefore depends on the temporal granularity
required by the downstream QA task rather than on a universally optimal
propagation policy.
\newpage
\section{Implementation Details}
\label{sec:implementation_details}

We evaluate all external baselines through their official implementations,
adapted to the same long-horizon conversational-memory benchmark interface.
This interface standardizes the visible memory prefix, query set, final evidence
budget, answer-generation prompt, and judging protocol across methods. Unless
otherwise stated, each method returns at most \(K=10\) evidence items per query.
For GPT-backed runs, LLM-dependent memory construction and answer generation use
\texttt{gpt-5.4-mini}. For Qwen-backed runs, they use
\texttt{qwen3.5-flash} with thinking disabled. All embedding-dependent
components use \texttt{text-embedding-3-small}. The judge is fixed to
\texttt{gpt-4o-mini} across methods.

\paragraph{BM25.}
BM25 is the lexical retrieval-only baseline. It indexes the same exchange-level
memory units as the other baselines and ranks visible memories with BM25 using
\(k_1=1.5\) and \(b=0.75\). It does not use LLM-based construction, embedding
retrieval, memory updates, graph construction, or hierarchical summarization.
The retrieved evidence budget is fixed to \(K=10\).

\paragraph{Dense retrieval.}
Dense retrieval is the flat embedding baseline. It embeds both the query and
each visible exchange-level memory unit with \texttt{text-embedding-3-small}
and ranks memory units by cosine similarity. The method does not build an
explicit memory structure and does not use an LLM during memory construction.
It returns the top \(K=10\) exchange-level memory units to the shared
answer-generation prompt.

\paragraph{RAPTOR.}
For RAPTOR~\citep{raptor2024}, we use the official recursive
clustering-and-summarization interface. The initial leaves are the benchmark
memory units. Node embeddings use \texttt{text-embedding-3-small}; cluster
summaries are generated by the LLM backbone used in the corresponding
experimental regime. The RAPTOR configuration uses reduction dimension \(3\),
cluster threshold \(0.5\), and maximum summary input length \(3500\) tokens.
The final retrieved evidence budget is \(K=10\).

\paragraph{MemTree.}
For MemTree~\citep{memtree2024}, we use the official online semantic-tree
interface. Memory units are inserted incrementally using the method's semantic
routing rule. The configuration sets the base insertion threshold to \(0.45\),
the threshold growth rate to \(1.0\), maximum depth to \(8\), and enables leaf
expansion. Internal aggregation uses the LLM backbone of the corresponding run,
and embeddings use \texttt{text-embedding-3-small}. Retrieval uses the
collapsed-tree mode with retrieval threshold \(0.10\), returning the top
\(K=10\) nodes.

\paragraph{A-MEM.}
For A-MEM~\citep{amem2025}, we use the official note-based memory interface.
Each memory unit is converted into a structured memory note with LLM-generated
JSON metadata. The text used for note embedding concatenates the note content,
context, keywords, and tags. The configuration uses \(2\) nearest neighbors for
note linking, enables memory evolution, sets the evolution threshold to \(100\),
and uses link-score threshold \(0.75\). Retrieval uses A-MEM's
semantic-plus-links mode and returns the top \(K=10\) memory notes.

\paragraph{Mem0.}
For Mem0~\citep{mem02025}, we use the official base Mem0 memory interface. We
do not report the graph-enhanced Mem0 variant in the main comparison. The
configuration uses additive single-pass memory extraction, normalized-text
MD5 hashing for deduplication, and entity linking. LLM-dependent extraction uses
the LLM backbone of the corresponding run, and memory embeddings use
\texttt{text-embedding-3-small}. Retrieval combines semantic, BM25, and entity
signals with configured weights \(0.70\), \(0.20\), and \(0.10\), respectively,
and uses a minimum overfetch of \(60\). The final evidence budget is \(K=10\).

\paragraph{HippoRAG.}
For HippoRAG~\citep{hipporag2024}, we use the official graph-retrieval
interface. LLM-dependent information extraction, query processing, and fact
filtering use the backbone of the corresponding experimental regime, while
embedding-dependent components use \texttt{text-embedding-3-small}. The
configuration sets the graph-retrieval damping factor to \(0.5\), synonymy
threshold to \(0.8\), query-linking top-\(k\) to \(5\), maximum filtered facts
to \(4\), and passage-node weight to \(0.05\). The top \(K=10\) retrieved
passages are passed to the shared answer-generation prompt.

\paragraph{Controlled deviations from the original baseline settings.}
The intentional deviations from some original baseline papers are the shared
benchmark substitutions required for controlled comparison: the LLM backbone,
embedding model, visible-memory interface, final evidence budget, answer prompt,
and judge are fixed across methods. In particular, all LLM-dependent baseline
components use either \texttt{gpt-5.4-mini} or \texttt{qwen3.5-flash} according
to the experimental regime, all embedding-dependent components use
\texttt{text-embedding-3-small}, and all reported runs use final evidence budget
\(K=10\). All other method-specific parameters follow the official
implementation interface and defaults, with the explicit configuration values
reported above.
\newpage
\section{Non-Temporal Similarity-Clustering Baseline}
\label{sec:appendix-nontemporal-baseline}

To isolate the role of temporal construction, we implement a non-temporal
similarity-clustering tree as a controlled baseline. This baseline uses the same
memory units, annotation prompts, embedding model, retrieval scorer, propagation
settings, and evidence budget as \ours{}. Its only intended difference from
\ours{} is the online update rule: instead of inserting each new utterance
through the rightmost temporal frontier, it assigns the utterance to the most
similar existing semantic cluster.

At time \(t\), the baseline maintains a clustering tree over the observed
utterance prefix \(X_t=(x_1,\ldots,x_t)\). Each utterance \(x_i\) is represented
as a leaf node. The baseline is initialized by creating the first lowest-level
cluster from the first leaf utterance. Let \(\mathcal C_t\) denote the set of
current lowest-level internal cluster nodes, which serve as the assignment
candidates for the next utterance. For each cluster \(c\in\mathcal C_t\), let
\(L(c)\) denote the set of leaf utterances under \(c\). The cluster is
represented by the mean embedding of its assigned utterances,
\[
\mu(c)
=
\frac{1}{|L(c)|}
\sum_{x_i\in L(c)}e(x_i),
\]
where \(e(\cdot)\) is the same embedding function used by the dense retrieval
pipeline.

When a new utterance \(x_{t+1}\) arrives, the baseline compares it against all
current lowest-level clusters and identifies the nearest cluster,
\[
c^\star
=
\arg\max_{c\in\mathcal C_t}
\cos\bigl(e(x_{t+1}),\mu(c)\bigr).
\]
If
\[
\cos\bigl(e(x_{t+1}),\mu(c^\star)\bigr)\ge \tau_{\mathrm{nt}},
\qquad
\tau_{\mathrm{nt}}=0.45,
\]
the new utterance is inserted as a new leaf child of \(c^\star\). Otherwise, the
baseline creates a new lowest-level internal cluster containing only the new
leaf. Thus, the number of active clusters is not fixed in advance, and new
clusters are created online when the incoming utterance is not sufficiently
similar to any existing cluster. Existing leaves are not reassigned during this
online update. After insertion, the embedding of the affected cluster and the
embeddings of its ancestors are recomputed, and their node annotations are
regenerated using the same summarization prompts as \ours{}.

This baseline is non-temporal because cluster membership is determined by
semantic similarity rather than by utterance order. Consequently, utterances
from distant parts of a conversation may be grouped under the same internal
node, even when they do not form a contiguous conversational episode. This
makes the baseline useful for testing the structural hypothesis of the paper:
whether a temporally ordered hierarchy provides a better memory state than a
semantic clustering tree under matched retrieval conditions. \Autoref{tab:appendix-nontemporal-control}
summarizes the main differences between this non-temporal baseline and \ours{}.

\begin{table}[H]
\centering
\small
\setlength{\tabcolsep}{4pt}
\renewcommand{\arraystretch}{1.08}
\begin{tabularx}{\linewidth}{@{}lXX@{}}
\toprule
\textbf{Component} & \textbf{\ours{}} & \textbf{Non-temporal baseline} \\
\midrule
Memory unit & Individual utterance & Same \\
Construction candidates & Rightmost temporal frontier & Current lowest-level semantic clusters \(\mathcal C_t\) \\
Assignment criterion & Frontier compatibility with the incoming utterance & Cosine similarity to cluster mean embeddings with threshold \(\tau_{\mathrm{nt}}=0.45\) \\
New-cluster creation & New temporal branch when no frontier node is compatible & New semantic cluster when nearest-cluster similarity is below \(\tau_{\mathrm{nt}}\) \\
Temporal constraint & Every internal node covers a contiguous span & No contiguity constraint \\
Leaf reassignment & Existing leaves are never reordered & Existing leaves are not reassigned \\
Annotation model & Same LLM summarization prompts & Same prompts \\
Retrieval & Same local scorer, propagation grid, and evidence budget & Same settings \\
\bottomrule
\end{tabularx}
\captionsetup{skip=10pt}
\caption{Controlled comparison between temporal Segment Tree construction and the non-temporal similarity-clustering baseline. The baseline is designed to test whether preserving chronological contiguity during construction improves the memory representation, while keeping retrieval and annotation choices fixed.}
\label{tab:appendix-nontemporal-control}
\end{table}
\newpage


\definecolor{promptpurplebg}{RGB}{248,244,255} \definecolor{promptpurpleframe}{RGB}{190,165,230} \definecolor{promptpurpletitle}{RGB}{82,45,125} \newtcblisting{promptbox}[1][]{ enhanced, breakable, colback=promptpurplebg, colframe=promptpurpleframe, coltitle=promptpurpletitle, fonttitle=\bfseries\small, fontupper=\footnotesize\ttfamily, boxrule=0.6pt, arc=2pt, left=6pt, right=6pt, top=5pt, bottom=5pt, listing only, listing options={ basicstyle=\footnotesize\ttfamily, breaklines=true, breakatwhitespace=true, columns=fullflexible, keepspaces=true, showstringspaces=false }, #1 }

\section{Prompt Templates}
\label{sec:appendix-prompts}

\subsection{\ours{} Construction Prompts}
\label{sec:appendix-segtreemem-prompts}

\ours{} uses LLMs only for the LLM-based construction variants and for
internal-node annotation. Retrieval and dataset loading do not invoke LLMs.
Specifically, the implementation uses three prompt templates: pointwise LLM
compatibility, batch LLM compatibility, and internal-node annotation.

\paragraph{Pointwise LLM compatibility.}
The pointwise construction variant evaluates one admissible rightmost-frontier
candidate at a time. The model decides whether the incoming subtree should be
merged into the candidate or whether construction should continue upward along
the frontier.

\begin{promptbox}[title={System prompt}]
You evaluate a single rightmost-frontier candidate for online segment-
tree memory construction.

Return exactly MERGE or SPLIT.

MERGE when the incoming subtree is semantically compatible with the
frontier node and should be appended as its rightmost child. SPLIT when
the incoming subtree should not merge with this candidate and should
continue upward.
\end{promptbox}

\begin{promptbox}[title={User prompt}]
Frontier node interval: [{frontier_node.left}, {frontier_node.right}]
Frontier node content:
{rendered frontier node}

Incoming subtree interval: [{current.left}, {current.right}]
Incoming subtree content:
{rendered incoming subtree}

Decision:
\end{promptbox}

\paragraph{Batch LLM compatibility.}
The batch construction variant evaluates the admissible rightmost frontier in a
single LLM call. The frontier candidates are ordered from lower to higher level.
If there are \(m\) admissible candidates, the valid merge labels are dynamically
generated as \texttt{MERGE\_1}, \(\ldots\), \texttt{MERGE\_m}. The model returns
exactly one merge label or \texttt{SPLIT}. A label \texttt{MERGE\_i} selects
candidate \(i\) as the attachment node, while \texttt{SPLIT} indicates that no
supplied frontier candidate is compatible with the incoming subtree.

\begin{promptbox}[title={System prompt}]
You evaluate the ordered rightmost frontier for online segment-tree
memory construction.

Return exactly one label: MERGE_1, MERGE_2, ..., MERGE_m, or SPLIT.

MERGE_i selects the single frontier candidate i whose node is
semantically compatible with the incoming subtree and should receive it
as the rightmost child. SPLIT means no supplied frontier candidate is
compatible.
\end{promptbox}

\begin{promptbox}[title={User prompt}]
Incoming subtree interval: [{incoming.left}, {incoming.right}]
Incoming subtree content:
{rendered incoming subtree}

Ordered frontier candidates, from lower to higher level:
Candidate 1 -> MERGE_1
Interval: [{candidate_1.left}, {candidate_1.right}]
Content:
{rendered candidate 1}

Candidate 2 -> MERGE_2
Interval: [{candidate_2.left}, {candidate_2.right}]
Content:
{rendered candidate 2}

... repeated for all candidates ...

Decision:
\end{promptbox}

\paragraph{Internal-node annotation.}
After construction creates or updates an internal node, \ours{} regenerates the
node annotation from its ordered child segments. The child segments are rendered
in chronological order. The annotation prompt is shared across construction
variants.

\begin{promptbox}[title={System prompt}]
You maintain a temporally ordered segment-tree memory. Write a concise
annotation for one internal node from its ordered child segments. Return
only the annotation text.
\end{promptbox}

\begin{promptbox}[title={User prompt}]
Node interval: [{node.left}, {node.right}]

Ordered child segments:
- interval=[{child.left},{child.right}] id={child.node_id}
{rendered child}

... repeated for each ordered child ...

Annotation:
\end{promptbox}
\subsection{LoCoMo Prompts}
\label{sec:appendix-prompts-locomo}

For LoCoMo, retrieved evidence is inserted into the following prompt
before answer generation.

\begin{promptbox}[title={Answer generaiton}]
You are a helpful assistant answering questions about a long-term conversation
between two people. Use the retrieved memory excerpts to answer the question
accurately and concisely.

Memories:
{evidence_text}

Question: {question}

Answer:
\end{promptbox}

We evaluate generated LoCoMo answers using the following rubric. The
judge receives the question, question category, gold answer, and
candidate answer, and returns an integer score from 0 to 3.

\begin{promptbox}[title={LLM judge prompt}]
Your task is to evaluate the quality of a candidate answer to a conversational
memory question.

You will be given:
1. The question asked
2. The question category (1=single-hop, 2=temporal, 3=open-ended,
   4=multi-hop, 5=adversarial)
3. The gold answer (or adversarial answer for category 5)
4. The candidate answer to evaluate

Scoring criteria:
Score 0: The candidate answer is incorrect or contradicts the gold answer. For
category 5, the candidate provides the adversarial (wrong) answer as if it were
true.
Score 1: The candidate answer is vague or generic, not using specific information
from the conversation. For category 5, the candidate is uncertain but doesn't
clearly reject the premise.
Score 2: The candidate answer is partially correct, capturing some but not all
key information from the gold answer. For category 5, the candidate correctly
identifies that the information may not be available.
Score 3: The candidate answer is fully correct and captures the key information
from the gold answer. For category 5, the candidate clearly states the
information is not in the conversation.

Important:
- Focus on factual accuracy, not style or fluency.
- For temporal questions (category 2), dates and time references must be correct.
- For multi-hop questions (category 4), all reasoning steps must be present.
- For adversarial questions (category 5), the correct behavior is to recognize
  the question cannot be answered from the conversation.
\end{promptbox}

\subsection{LongMemEval-MAB Prompts}
\label{sec:appendix-prompts-longmemeval}

For LongMemEval-MAB, retrieved evidence is inserted into the following
answer-generation prompt.

\begin{promptbox}[title={Answer generation}]
You are a helpful assistant that answers questions based on your memory of
previous conversations. Use the retrieved memory excerpts to answer accurately
and concisely.

Memories:
{evidence_text}

Question: {question}

Answer:
\end{promptbox}

\subsection{RealMem Prompts}
\label{sec:appendix-prompts-realmem}

For RealMem, we use the official answer-generation prompt from the
RealMemBench evaluation code.

\begin{promptbox}[title={Answer generation}]
You are a personal AI assistant that helps the user with some long-term tasks.
Please respond to the user's latest message based on the reference memory.

Memories:{evidence_text}

Query: {question}

Response:
\end{promptbox}

RealMem evaluates answer quality by checking consistency between the
candidate answer and the user-related memory.

\begin{promptbox}[title={LLM judge prompt}]
Your task is to evaluate the consistency between the [candidate answer] and the
[user-related memory].

You will be given four pieces of information:
1. The user's current query
2. The user-related memory, representing the latest valid user state
3. A reference answer based on the relevant memory
4. The candidate answer to be evaluated

Please follow these rules during evaluation:
- Focus only on whether "facts, constraints, preferences, and confirmed states"
  are correctly used
- Do NOT evaluate language style, tone, politeness, empathy, or fluency
- Do NOT give a high score just because the answer "sounds reasonable"
- The reference answer is only to help understand how relevant memory should
  ideally be used. A candidate answer does not need to exactly match the
  reference answer to receive a full score

Scoring criteria:
Score 0: Poor -- the candidate answer conflicts with the user-related memory
Score 1: Fair -- the candidate answer does not conflict with the relevant memory
but is generic and not based on user memory
Score 2: Good -- the candidate answer uses part of the user-related memory
Score 3: Very good -- the candidate answer uses all of the user-related memory

Output format:

{
    "score": int,
    "reason": str
}
\end{promptbox}

\section{Limitations and Broader Impacts}
\label{sec:appendix-limitations}

\paragraph{Limitations.}
Agentic memory systems remain sensitive to the quality of their stored representations, retrieval mechanisms, and summarization procedures. Errors introduced during memory construction can persist over long horizons, causing agents to retrieve incomplete, outdated, or misleading context. More broadly, current memory systems still lack robust guarantees for reliability, privacy preservation, and under noisy, adversarial, or rapidly changing user histories.

\paragraph{Broader Impacts.}
Long-term agentic memory has the potential to make AI assistants more useful by supporting continuity across extended interactions, personalized help, and better handling of evolving user goals. Persistent memory also creates risks around privacy, data retention, and over-personalization. Users should remain in control over what is stored, retrieved, edited, and forgotten, and careful analysis should be done to ensure memory improves user outcomes without amplifying harmful or sensitive inferences.
\newpage
\section*{NeurIPS Paper Checklist}
\label{sec:checklist}

\begin{enumerate}

\item {\bf Claims}
    \item[] Question: Do the main claims made in the abstract and introduction accurately reflect the paper's contributions and scope?
    \item[] Answer: \answerYes{}
    \item[] Justification: Our abstract and introduction accurately summarize our main contributions: a temporally ordered segment-tree memory representation, online rightmost-frontier construction, and structure-aware relevance propagation. We evaluate the empirical claims across LoCoMo, LongMemEval-MAB, and RealMem in \Autoref{sec:experiments}.

\item {\bf Limitations}
    \item[] Question: Does the paper discuss the limitations of the work performed by the authors?
    \item[] Answer: \answerYes{}
    \item[] Justification: We discuss limitations in \appref{sec:appendix-adversarial} and \appref{sec:appendix-qa-cases}, including maximally topic-switching conversations, long-range topical recurrence, and cases where internal node summaries miss exact details. We also discuss cases where propagation can dilute local evidence and broader limitations of long-term memory systems in \appref{sec:appendix-limitations}.
    
\item {\bf Theory assumptions and proofs}
    \item[] Question: For each theoretical result, does the paper provide the full set of assumptions and a complete (and correct) proof?
    \item[] Answer: \answerNA{}
    \item[] Justification: We formalize the memory representation, online update rule, and retrieval operator, but we do not present theoretical theorems requiring formal proofs.

\item {\bf Experimental result reproducibility}
    \item[] Question: Does the paper fully disclose all the information needed to reproduce the main experimental results of the paper to the extent that it affects the main claims and/or conclusions of the paper (regardless of whether the code and data are provided or not)?
    \item[] Answer: \answerYes{}
    \item[] Justification: We provide the information needed to reproduce the main experimental results. Sections~5.2--5.4 specify the datasets, baselines, evaluation protocol, model choices, metrics, and main experimental settings. \appref{sec:implementation_details} and \appref{sec:appendix-prompts} provide implementation details and prompt templates for construction, retrieval, answer generation, and evaluation.

\item {\bf Open access to data and code}
    \item[] Question: Does the paper provide open access to the data and code, with sufficient instructions to faithfully reproduce the main experimental results, as described in supplemental material?
    \item[] Answer: \answerYes{}
    \item[] Justification: We provide anonymized code in the supplemental material, including instructions for reproducing the main experiments. We use publicly available benchmark datasets and document the dataset choices, baselines, model settings, retrieval configuration, and prompt templates in \Autoref{sec:experiments} and \appref{sec:implementation_details} and \appref{sec:appendix-prompts}.
   
\item {\bf Experimental setting/details}
    \item[] Question: Does the paper specify all the training and test details (e.g., data splits, hyperparameters, how they were chosen, type of optimizer) necessary to understand the results?
    \item[] Answer: \answerYes{}
    \item[] Justification: We specify the datasets, baselines, evaluation metrics, answer-generation models, judge model, and main experimental configuration in \Autoref{sec:experiments}. We provide additional details on construction, retrieval, answer generation, and evaluation prompts in \appref{sec:implementation_details} and \appref{sec:appendix-prompts}.
    
\item {\bf Experiment statistical significance}
    \item[] Question: Does the paper report error bars suitably and correctly defined or other appropriate information about the statistical significance of the experiments?
    \item[] Answer: \answerYes{}
    \item[] Justification: We report 95\% confidence intervals for the main LLM-judge accuracy and token-level F1 results in \Autoref{tab:main_results}. We compute these intervals over evaluation examples, so they quantify uncertainty due to finite benchmark size rather than variation across model training runs.

\item {\bf Experiments compute resources}
    \item[] Question: For each experiment, does the paper provide sufficient information on the computer resources (type of compute workers, memory, time of execution) needed to reproduce the experiments?
    \item[] Answer: \answerYes{}
    \item[] Justification: We describe the compute setting in \appref{sec:implementation_details}. Our experiments use pretrained API models and local orchestration rather than model training or finetuning. We identify the main computational costs, including embedding memory nodes, LLM-based node annotation, LLM-based compatibility scoring, answer generation, and LLM-as-a-judge evaluation.

\item {\bf Code of ethics}
    \item[] Question: Does the research conducted in the paper conform, in every respect, with the NeurIPS Code of Ethics \url{https://neurips.cc/public/EthicsGuidelines}?
    \item[] Answer: \answerYes{}
    \item[] Justification: We use public or benchmark conversational-memory datasets and evaluate retrieval and answer generation methods. We have reviewed the NeurIPS Code of Ethics and do not identify deviations from it.

\item {\bf Broader impacts}
    \item[] Question: Does the paper discuss both potential positive societal impacts and negative societal impacts of the work performed?
    \item[] Answer: \answerYes{}
    \item[] Justification: We discuss both positive and negative broader impacts in \appref{sec:appendix-limitations}. We note that long-term memory can improve continuity, personalization, and support for evolving user goals, while also creating risks around privacy, data retention, over-personalization, and harmful or sensitive inferences.

\item {\bf Safeguards}
    \item[] Question: Does the paper describe safeguards that have been put in place for responsible release of data or models that have a high risk for misuse (e.g., pre-trained language models, image generators, or scraped datasets)?
    \item[] Answer: \answerNA{}
    \item[] Justification: We do not release a pretrained model, scraped dataset, or other high-risk artifact. We evaluate memory construction and retrieval methods on existing benchmarks.

\item {\bf Licenses for existing assets}
    \item[] Question: Are the creators or original owners of assets (e.g., code, data, models), used in the paper, properly credited and are the license and terms of use explicitly mentioned and properly respected?
    \item[] Answer: \answerYes{}
    \item[] Justification: We cite the original sources for the datasets, baselines, pretrained models, and evaluation assets used in our experiments. We do not redistribute the original benchmark datasets or pretrained models; users should obtain those assets from their original sources and comply with the corresponding licenses and terms of use. Our supplemental code documents dependencies and reproduction instructions.

\item {\bf New assets}
    \item[] Question: Are new assets introduced in the paper well documented and is the documentation provided alongside the assets?
    \item[] Answer: \answerYes{}
    \item[] Justification: We release anonymized supplemental code as a new asset. We do not release a new dataset or pretrained model.

\item {\bf Crowdsourcing and research with human subjects}
    \item[] Question: For crowdsourcing experiments and research with human subjects, does the paper include the full text of instructions given to participants and screenshots, if applicable, as well as details about compensation (if any)? 
    \item[] Answer: \answerNA{} 
    \item[] Justification: Not applicable

\item {\bf Institutional review board (IRB) approvals or equivalent for research with human subjects}
    \item[] Question: Does the paper describe potential risks incurred by study participants, whether such risks were disclosed to the subjects, and whether Institutional Review Board (IRB) approvals (or an equivalent approval/review based on the requirements of your country or institution) were obtained?
    \item[] Answer: \answerNA{} 
    \item[] Justification: Not applicable

\item {\bf Declaration of LLM usage}
    \item[] Question: Does the paper describe the usage of LLMs if it is an important, original, or non-standard component of the core methods in this research? Note that if the LLM is used only for writing, editing, or formatting purposes and does \emph{not} impact the core methodology, scientific rigor, or originality of the research, declaration is not required.
   \item[] Answer: \answerYes{}
    \item[] Justification: We use LLMs as core components of both our method and evaluation. In the method, we use LLMs to generate node annotations and, in the LLM-based construction variants, to make compatibility judgments during online tree construction. In the experiments, we use LLMs for answer generation and LLM-as-a-judge evaluation. We describe these uses in \Autoref{sec:method} and \Autoref{sec:experiments} and provide the corresponding prompts in \appref{sec:appendix-prompts}.

\end{enumerate}

\end{document}